\documentclass[10pt,twocolumn,letterpaper]{article}

\usepackage{cvpr}
\usepackage{times}
\usepackage{epsfig}
\usepackage{graphicx}
\usepackage{amsmath}
\usepackage{amssymb}
\usepackage{subcaption}

\usepackage[utf8]{inputenc} % allow utf-8 input
\usepackage[T1]{fontenc}    % use 8-bit T1 fonts
\usepackage{url}            % simple URL typesetting
\usepackage{booktabs}       % professional-quality tables
\usepackage{amsfonts}       % blackboard math symbols
\usepackage{nicefrac}       % compact symbols for 1/2, etc.
\usepackage{microtype}      % microtypography

\usepackage{caption}
\usepackage{multirow}
\usepackage{xspace}
\usepackage{mathtools}
\usepackage{wasysym}

\usepackage{marvosym}

\usepackage[table,xcdraw]{xcolor}
\usepackage{pdfcomment}
\usepackage{lipsum}
\usepackage{algorithm}
\usepackage[noend]{algpseudocode}
\usepackage{wrapfig}

\usepackage{hyperref}
\hypersetup{breaklinks=true,colorlinks}

\cvprfinalcopy % *** Uncomment this line for the final submission

\def\cvprPaperID{2589} % *** Enter the CVPR Paper ID here

\ifcvprfinal\fi

\def\cp{{\mathbf{cp}}}

\def\bp{{\bf p}}
\makeatother

\title{\vspace{-3.5mm}Fast Interactive Object Annotation with Curve-GCN}

\author{Huan Ling \And Jun Gao \And  Amlan Kar \And  Wenzheng Chen \And Sanja Fidler
}

\author{Huan Ling$^{1,2}$\thanks{authors contributed equally}\hspace{1cm} Jun Gao$^{1,2}$\footnotemark[1] \hspace{1cm} Amlan Kar$^{1,2}$ \hspace{1cm} Wenzheng Chen $^{1,2}$ \hspace{1cm} Sanja Fidler$^{1,2,3}$\\
$^1$University of Toronto \hspace{1em} $^2$Vector Institute \hspace{1em}  \hspace{1em} $^3$NVIDIA\\
{\tt\small \{linghuan, jungao, amlan, wenzheng, fidler\}@cs.toronto.edu}}

\begin{document}

\maketitle
\begin{abstract}
Manually labeling objects by tracing their boundaries is a laborious process. In~\cite{polyrnn,polyrnnpp}, the authors proposed Polygon-RNN that produces polygonal annotations in a recurrent manner using a CNN-RNN architecture, allowing interactive correction via humans-in-the-loop. We propose a new framework that alleviates the sequential nature of Polygon-RNN, by predicting all vertices simultaneously using a Graph Convolutional Network (GCN). Our model is trained end-to-end. %, and runs in real time. 
It supports object annotation by either polygons or splines, facilitating labeling efficiency for both line-based and curved objects. %In particular, we parametrize the shape of an object with a spline, and iteratively regress to the locations of its control points using a Graph Convolutional Network (GCN).
%Our model, referred to as Spline-GCN, is trained using a differentiable loss function that optimizes foewr accuracy.
We show that Curve-GCN outperforms all existing approaches in automatic mode, including the powerful PSP-DeepLab~\cite{ChenPKMY18,dextr} and is significantly more efficient in interactive mode than Polygon-RNN++. Our model runs at 29.3ms in automatic, and 2.6ms in interactive mode, making it 10x and 100x faster than Polygon-RNN++.
\end{abstract}

%!TEX root = top.tex
\vspace{-4mm}
\section{Introduction}
\label{sec:intro}
% \vspace{-2mm}

Object instance segmentation is the problem of outlining all objects of a given class in an image, a task that has been receiving increased attention in the past few years~\cite{maskrcnn,ZhangCVPR16,SGN17,DWT17,liu2018path}. %It facilitates many downs-stream robotic tasks such as grasping~\cite{pathakCVPRW18}, future prediction~\cite{future}, medical image analysis such as cell classification~\cite{ValenKLMQDMTAC16}, as well as tasks like image editing~\cite{sss} or image generation~\cite{wang2018pix2pixHD}. For these applications, it is essential to identify object boundaries with high accuracy. 
%Several recent deep models have shown impressive progress in this domain~\cite{maskrcnn,ZhangCVPR16,SGN17,DWT17,liu2018path}. 
%Arguably, learned models can only be as good as the labeled data they are trained on, making it extremely important to collect datasets with very precise annotations. 
Current approaches are all data hungry, and benefit from large annotated datasets for training. 
However, manually tracing object boundaries is a laborious process, taking up to 40sec per object~\cite{polyrnnpp,ChenCVPR14}. To alleviate this problem, a number of interactive image segmentation techniques have been proposed~\cite{Rother2004SIGGRAPH,dextr,polyrnn,polyrnnpp}, speeding up annotation by a significant factor. We follow this line of work.

In DEXTR~\cite{dextr}, the authors build upon the Deeplab architecture~\cite{ChenPKMY18} by incorporating a simple encoding of human clicks in the form of heat maps. 
This is a pixel-wise approach, \ie it predicts a foreground-background label for each pixel. %Pixel-wise outputs are harder to correct by an annotator in that the edits need to be on the level of the pixels. 
DEXTR showed that by incorporating user clicks as a soft constraint, the model learns to interactively improve its prediction. Yet, since the approach is pixel-wise, the worst case scenario still requires many clicks.
%the annotator to click on many pixels. 

\begin{figure}[t!]
\vspace{-2mm}
\includegraphics[width=\linewidth,trim=0 0 0 0,clip]{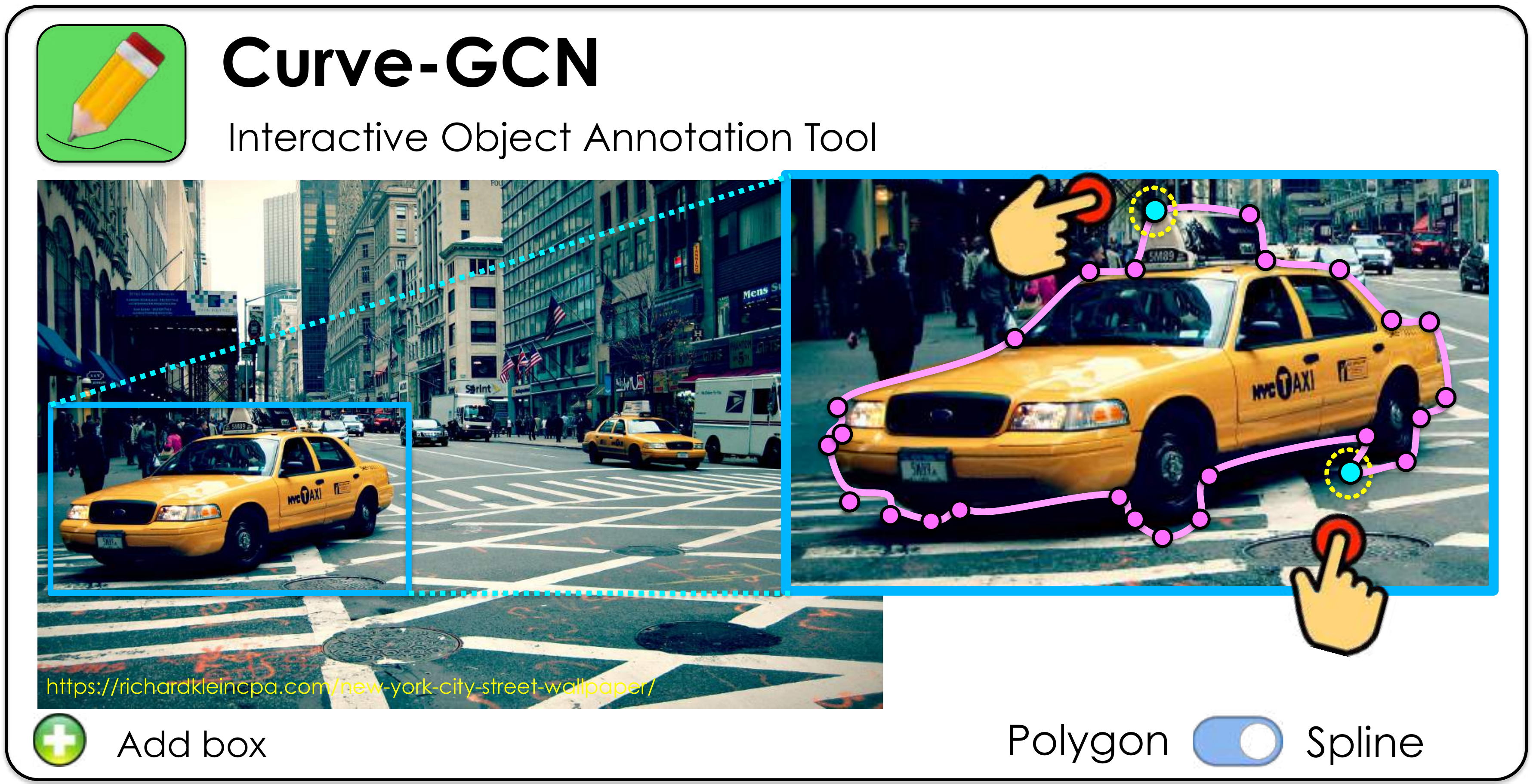} 
\vspace{-7mm}
\caption{\small We propose Curve-GCN for interactive object annotation. In contrast to Polygon-RNN~\cite{polyrnn,polyrnnpp}, our model parametrizes objects with either polygons or splines and is trained end-to-end at a high output resolution.} %Spline-GCN gives the annotator additional control over the radius in %which the change due to human edit may take effect.}% The model runs 10x faster than~\cite{polyrnnpp}, requiring 30ms to make a prediction, and 3ms to incorporate each human interaction.}
\label{fig:intro}
\vspace{-3mm}
\end{figure}

Polygon-RNN~\cite{polyrnn,polyrnnpp} frames human-in-the-loop annotation as a recurrent process, during which the model sequentially predicts vertices of a polygon. The annotator can intervene whenever an error occurs, by correcting the wrong vertex. The model continues its prediction by conditioning on the correction. Polygon-RNN was shown to produce annotations at human level of agreement with only a few clicks per object instance. The worst case scenario here is bounded by the number of polygon vertices, which for most objects ranges up to 30-40 points. However, the recurrent nature of the model limits scalability to more complex shapes, resulting in harder training and longer inference. Furthermore, the annotator is expected to correct mistakes in a sequential order, which is often challenging in practice. % (\ie, in a real annotation tool~\cite{polydemo}). 

In this paper, we frame object annotation as a regression problem, where the locations of all vertices are predicted simultaneously. We represent the object as a graph with a fixed topology, and perform prediction using a Graph Convolutional Network. We show how the model can be used and optimized for interactive annotation. %, allowing the annotator to correct \emph{any} vertex, and 
%also be able to control the influence of the correction.
%focus on local correction. %\JUN{remove control k issue.} 
Our framework further allows us to parametrize objects with either polygons or splines, adding additional flexibility and efficiency to the interactive annotation process. The proposed approach, which we refer to as Curve-GCN, is end-to-end differentiable, and runs in real time. 
%While polygons are good at outlining man-made objects, splines more efficiently handle curved objects using fewer control points. 
%parametrize object shapes with splines. Splines are an even more efficient representation than polygons since they are defined with a smaller number of control points. We propose a Spline-GCN, which annotates object boundaries by iteratively regressing to the positions of the control points using a Graph Convolutional Network. The model is trained using a differentiable rendering loss that optimizes overlap accuracy. We make our model interactive by allowing the annotator to correct any vertex, and train Spline-GCN to re-predict the neighboring vertices. This is in contrast to Polygon-RNN~\cite{polyrnn}, where the annotator is expected to correct in sequential order, which is sometimes challenging in practice (\ie, in a real annotation tool~\cite{polydemo}). 
%The annotator is given additional control on how far (how many neighbors) the change can take affect. 
We evaluate our Curve-GCN on the challenging Cityscapes dataset~\cite{cityscapes}, where we outperform Polygon-RNN++ and PSP-Deeplab/DEXTR in both automatic and interactive settings.  We also show that our model outperforms the baselines in cross-domain annotation, that is, a model trained on Cityscapes is used to annotate general scenes~\cite{ade20k}, aerial~\cite{rooftop}, and medical imagery~\cite{medical1,sstem}. %Furthermore, in automatic mode, our model is X times faster than Polygon-RNN++ , and Y times faster than DEXTR. When incorporating user clicks, our model runs at X speed of Polygon-RNN, and Y speed of DEXTR, achieving run-time of XXms per click. 
Code is available: {\footnotesize\url{https://github.com/fidler-lab/curve-gcn}}.
%!TEX root = top.tex
% \vspace{-3mm}
\section{Related Work}
\label{sec:related}
% \vspace{-2mm}

{\bf Pixel-wise methods.} Interactive object segmentation has typically been formulated as a pixel-wise foreground-background segmentation. Most of the early work relies on optimization by graph-cuts to solve an energy function that depends on various color and texture cues~\cite{Rother2004SIGGRAPH,Boykov2001ICCV,ChenCVPR14}. The user is required to draw a box around the object, and can interact with the method by placing additional scribbles on the foreground or background, until the object is carved out correctly. However, in ambiguous cases where object boundaries blend with background, these methods often require many clicks from the user~\cite{polyrnnpp}.

Recently, DEXTR~\cite{dextr} incorporated user clicks by stacking them as additional heatmap channels to image features, and exploited the powerful Deeplab architecture to perform user-guided segmentation. The annotator is expected to click on the four extreme points of the object, and if necessary, iteratively add clicks on the boundary to refine prediction.  
Our work differs from the above methods in that it directly predicts a polygon or spline around the object, and avoids pixel-labeling altogether. We show this to be a more efficient way to perform object instance segmentation, both in the automatic and in the interactive settings. 

{\bf Contour-based methods.} Another line of work to object segmentation aims to trace closed contours.
Oldest techniques are based on level sets~\cite{Caselles}, which find object boundaries via front propagation by
solving a corresponding partial differential equation. Several smoothing terms help the contour evolution to be well behaved, producing accurate and regularized boundaries. 
In~\cite{david19edges}, levelset evolution with carefully designed boundary prediction was used to find accurate object boundaries from coarse annotations. This speeds up annotation since the annotators are only required to perform very coarse labeling.~\cite{zian19levelset} combines CNN feature learning with level set optimization in an end-to-end fashion, and exploits extreme points as a form of user interaction.
While most level set-based methods were not interactive,~\cite{levelsetCremers} proposed to incorporate user clicks into the energy function.  Recently,~\cite{marcos2018learning} proposed a structure prediction framework to learn CNN features jointly with the active contour parameters by optimizing an approximate IoU loss.  Rather than relying on the regularized contour evolution which may lead to overly smooth predictions, our approach learns to perform inference using a GCN. We further tackle the human-in-the-loop scenario, not addressed in~\cite{marcos2018learning}. 

Intelligent Scissors~\cite{Mortensen} is a technique that allows the user to place ``seeds'' along the boundary and finds the minimal cost contour starting from the last seed up to the mouse cursor, by tracing along the object's boundary. In the case of error, the user is required to place more seeds. 
%The method needed many interactions from the user, however, there may be potential in marrying this approach with deep learning. 

Polygon-RNN~\cite{polyrnn,polyrnnpp} adopted a similar idea of sequentially tracing a boundary, by exploiting a CNN-RNN architecture. Specifically, the RNN predicts a polygon by outputting one vertex at a time. However, the recurrent structure limits the scaleability with respect to the number of vertices, and also results in slower inference times.
%Prediction at each time-step is framed as a classification task, \ie choosing the right location in a small image grid.  The model was trained using cross-entropy at each time step, while Polygon-RNN++~\cite{polyrnn} improved upon this using Reinforcement Learning. 
%Polygon-RNN++ additionally proposed a refinement step, that takes the predicted polygon and predicts a finer set of vertices at a higher resolution using a Graph Gated Neural Network. 
Our work is a conceptual departure from Polygon-RNN in that we frame object annotation as a regression problem, where the locations of all vertices of the polygon are predicted simultaneously. The key advantages of our approach is that our model is significantly faster, and can be trained end-to-end using a differentiable loss function. Furthermore, our model is designed to be invariant to order, thus allowing the annotator to correct \emph{any} vertex, and further control the influence of the correction. 

Our approach shares similarities with Pixel2Mesh~\cite{pixel2mesh}, which predicts 3D meshes of objects from single images. We exploit their iterative inference regime, but propose a different parametrization based on splines and a loss function better suited for our (2D annotation) task. Moreover, we tackle the human-in-the-loop scenario, not addressed in~\cite{pixel2mesh}. 

%We take inspiration from this approach, but propose a model that is end-to-end differentiable. We achieve this by fixing the topology of a spline (\ie, the number of control points), and use a differentiable rendering loss to train the model. We additionally follow Pixel2Mesh~\cite{pixel2mesh}, and perform iterative coarse-to-fine prediction. 

Splines have been used to parametrize shapes in older work on active shape models~\cite{tan2014active}. However, these models are not end-to-end, while also requiring a dataset of aligned shapes to compute the PCA basis. Furthermore, interactivity comes from the fact that the prediction is a spline that the user can modify. In our approach, every modification leads to re-prediction, resulting in much faster interactive annotation. 
%!TEX root = top.tex
% \vspace{-5mm}
\section{Object Annotation via Curve-GCN}
\label{sec:method}

%We now introduce our framework, which annotates object instances with either polygons or (closed) splines. %Polygons may be a more efficient representation for man-made like objects like buildings, however, most of the natural objects are curved.
%Polygons only represent the geometry of a small fraction of objects found in the real world efficiently. 
In order to approximate a curved contour outlining an object, one can either draw a polygon or a spline. Splines are a more efficient form of representation as they allow precise approximation of the shape with fewer control points. Our framework is designed to enable both a polygon and a spline representation of an object contour.

We follow the typical labeling scenario where we assume that the annotator has selected the object of interest by placing a bounding box around it~\cite{polyrnn,polyrnnpp}. We crop the image around this box and frame object annotation in the crop as a regression problem; to predict the locations of all control points (vertices) simultaneously, from an initialization with a fixed topology.
%In particular, we assume that the object can be represented using a polygon or spline with a constant number of $N$ control points.
%This is a conceptual departure from Polygon-RNN~\cite{} which produces the vertices sequentially. The significant advantages of our approach is that the model can be trained end-to-end using a differentiable loss function. Furthermore, our model is designed to invariant to order, thus allowing the annotator to correct \emph{any} vertex, and further control the locality of the correction.
We describe our model from representation to inference in Subsec.~\ref{sec:model}, and discuss training in Subsec~\ref{sec:training}. In Subsec.~\ref{sec:interactive}, we explain how our model can be used for human-in-the loop annotation, by formulating both inference as well as training in the interactive regime.

%In this section, we introduce our model, Spline-GCN. We first present the base model, Polygon-GCN, annotating objects with polygon following PolygonRNN~\cite{polyrnn}. We then extend the polygon annotation to spline curve and instantiate Spline-GCN in the section~\ref{sec:spline-gcn}. The loss function to train Spline-GCN is provided in the section~\ref{sec:loss}, and learning with incorporating human in the loop is introduced at the last section. Our model does forground-background instance segmentation when given a bounding box of region of interests. The full pipline is shown in Fig~\ref{fig:full_model}

\begin{figure*}[t!]
\vspace{-3mm}
\includegraphics[width=\linewidth,trim=0 0 0 0,clip]{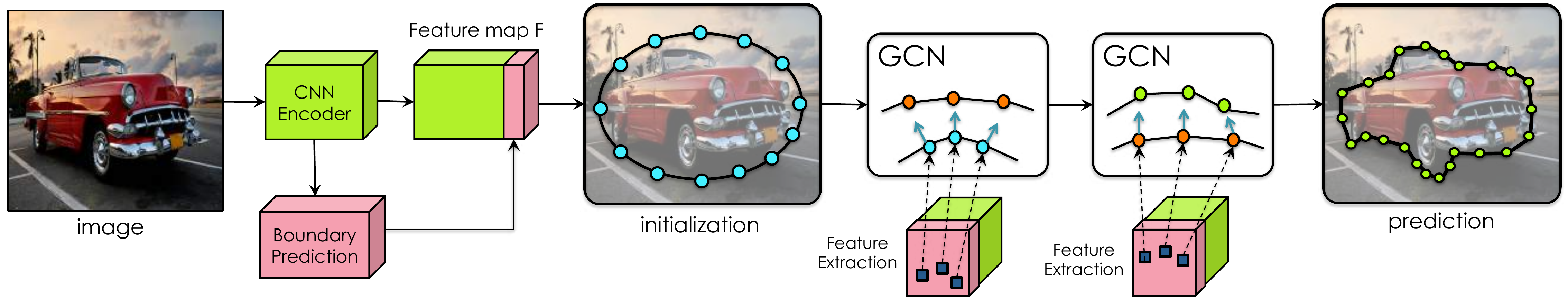}
\vspace{-8mm}
\caption{\footnotesize {\bf Curve-GCN}: We initialize $N$ control points (that form a closed curve) along a circle centered in the image crop with a diameter of $70\%$ of image height. We form a graph and propagate messages via a Graph Convolutional Network (GCN) to predict a location shift for each node. This is done iteratively (3 times in our work). At each iteration we extract a feature vector for each node from the CNN's features $F$, using a bilinear interpolation kernel.}
\label{fig:full_model}
\vspace{-3mm}
\end{figure*}

\subsection{Polygon/Spline-GCN}
\label{sec:model}
We assume our target object shapes can be well represented using $N$ control points, which are connected to form a cycle.  %This means that each object has the same topology, \ie each control point is connected to two other control points.
The induced shape is rendered by either connecting them with straight lines (thus forming a polygon), or higher order curves (forming a spline). We treat the location of each control point as a continuous random variable, and learn to predict these via a Graph Neural Network that takes image evidence as input. In~\cite{polyrnnpp}, the authors exploited Gated Graph Neural Networks (GGNN)~\cite{GGNN} as a polygon refinement step, in order to upscale the vertices output by the RNN to a higher resolution. In similar vein, Pixel2Mesh~\cite{pixel2mesh} exploited a Graph Convolutional Network (GCN) to predict vertex locations of a 3D mesh. The key difference between a GGNN and a GCN is in the graph information propagation; a GGNN shares propagation matrices through time akin to a gated recurrent unit (GRU), whereas a GCN has propagation steps implemented as unshared ``layers'', similar to a typical CNN architecture. We adopt the GCN in our model due to its higher capacity. Hence, we name our model, \emph{Curve-GCN}, which includes \emph{Polygon} or \emph{Spline-GCN}.
\vspace{-3mm}
\paragraph{Notation:} We initialize the nodes of the GCN to be at a static initial central position in the given image crop (Fig.~\ref{fig:full_model}). Our GCN predicts a location offset for each node, aiming to move the node correctly onto the object's boundary. Let $\cp_i=[x_i,y_i]^T$ denote the location of the $i$-th control point and $ V = \{\cp_0, \cp_1, \cdots, \cp_{N-1}\}$ be the set of all control points. We define the graph to be $G = (V, E)$, with $V$ defining the nodes and $E$ the edges in the graph. We form $E$ by connecting each vertex in $V$ with its four neighboring vertices. %, \ie, $\cp_i$ is linked with $\cp_{(i-2)\%N}, \cp_{(i-1)\%N}, \cp_{(i+1)\%N}, \cp_{(i+2)\%N}$. %This connectivity is for the sole purpose of information propagation in the GCN, and does not reflect the way the shape is being ``rendered''.
This graph structure defines how the information propagates in the GCN. Connecting 4-way allows faster exchange of information between the nodes in the graph. %, and is a choice made based on experimentation. Other connectivities are possible.
\vspace{-3mm}
\paragraph{Extracting Features:} Given a bounding box, we crop the corresponding area of the image and encode it using a CNN, the specific choice of which we defer to experiments. We denote the feature map obtained from the last convolutional layer of the CNN encoder applied on the image crop as $F_c$. In order to help the model see image boundaries, we supervise two additional  branches, \ie an edge branch and a vertex branch, on top of the CNN encoder's feature map $F_c$, both of which consist of one 3 $\times$ 3 convolutional layer and one fully-connected layer. These branches are trained to predict the probability of existence of an object edge/vertex on a 28 $\times$ 28 grid. We train these two branches with the binary cross entropy loss. The predicted edge and vertices outputs are concatenated with $F_c$, to create an augmented feature map $F$. The input feature for a node $\cp_i$ in the GCN is a concatenation of the node's current coordinates $(x_i, y_i)$, where top-left of the cropped images is $(0,0)$ and image length is $1$, and features extracted from the corresponding location in $F$:$f_{i}^{0} = \mathrm{concat}\{F(x_i,y_i) ,x_i, y_i\}$.
Here, $F(x_i,y_i)$ is computed using bilinear interpolation.
\vspace{-3mm}
\paragraph{GCN Model:} We utilize a multi-layer GCN. The graph propagation step for a node $\cp_i$ at layer $l$ is expressed as:
\begin{equation}
	f_{i}^{l+1} = w_0^l f_{i}^l + \sum_{\cp_j \in \mathcal{N}(\cp_i)} w_1^l f_j^l
\end{equation}
where $\mathcal{N}(\cp_i)$ denotes the nodes that are connected to $\cp_i$ in the graph, and $w_0^l, w_1^l$ are the weight matrices. Following~\cite{GCN, pixel2mesh}, we utilize a Graph-ResNet to propagate information between the nodes in the graph as a residual function. The propagation step in one full iteration at layer $l$ then takes the following form:\\[-5mm]
\begin{eqnarray}
	r_{i}^{l} &=& ReLU\big(w_0^{l} f_{i}^l + \sum_{\cp_j \in \mathcal{N}(\cp_i)} w_1^{l} f_j^l\big ) \\
	r_{i}^{l+1} &=& \tilde w_0^{l} r_{i}^l + \sum_{\cp_j \in \mathcal{N}(\cp_i)} \tilde w_1^{l}r_j^l\\
	f_{i}^{l+1} &=& ReLU(r_{i}^{l+1} + f_{i}^l),
\end{eqnarray}
where $w_0, w_1, \tilde w_0, \tilde w_1$ are weight matrices for the residual. On top of the last GCN layer, we apply a single fully connected layer to take the output feature and predict a relative location shift, $(\Delta x_i, \Delta y_i)$,  for each node, placing it into location $[x_i',y_i']=[x_i + \Delta x_i, y_i + \Delta y_i]$. We also perform iterative inference similar to the coarse-to-fine prediction in~\cite{pixel2mesh}. To be specific, the new node locations ${[x_i',y_i']}$ are used to re-extract features for the nodes, and another GCN predicts a new set of offsets using these features. This mimics the process of the initial polygon/spline iteratively ``walking'' towards the object's boundaries.

\vspace{-3mm}
\paragraph{Spline Parametrization:}
%\subsection{Spline-GCN}
%\label{sec:spline-gcn}
The choice of spline is important, particularly for the annotator's experience. The two most common splines, \ie the cubic Bezier spline and the uniform B-Spline~\cite{prautzsch2013bezier,gao2019deepspline}, are defined by control points which do not lie on the curve, which could potentially confuse an annotator that needs to make edits. Following~\cite{tan2014active}, we use the centripetal Catmull-Rom spline (CRS)~\cite{yuksel2011parameterization}, which has control points along the curve. We refer the reader to~\cite{yuksel2011parameterization} for a detailed visualization of different types of splines.

For a curve segment $\textbf{S}_i$ defined by control points $\cp_{i-1}$, $\cp_{i}$, $\cp_{i+1}$, $\cp_{i+2}$ and a knot sequence $t_{i-1},t_{i},t_{i+1},t_{i+2}$, the CRS is interpolated by:
\begin{eqnarray}
\label{eq:crs-spline}
	\textbf{S}_i = \tfrac{t_{i+1} - t}{t_{i+1} - t_i}L_{012} +  \tfrac{t - t_i}{t_{i+1} - t_i}L_{123}
\end{eqnarray}
where\\[-7mm]
\begin{eqnarray}
	L_{012} &=&\tfrac{t_{i+1} - t}{t_{i+1} - t_{i-1}}L_{01} +  \tfrac{t - t_{i-1}}{t_{i+1} - t_{i-1}}L_{12}\\
	L_{123} &=&\tfrac{t_{i+2} - t}{t_{i+2} - t_{i}}L_{12} +  \tfrac{t - t_{i}}{t_{i+2} - t_{i}}L_{23}\\
	L_{01} &=&\tfrac{t_{i} - t}{t_{i} - t_{i-1}}\cp_{i-1} +  \tfrac{t - t_{i-1}}{t_{i} - t_{i-1}}\cp_{i}\\
	L_{12} &=&\tfrac{t_{i+1} - t}{t_{i+1} - t_{i}}\cp_{i} +  \tfrac{t - t_{i}}{t_{i+1} - t_{i}}\cp_{i+1}\\
	L_{23} &=&\tfrac{t_{i+2} - t}{t_{i+2} - t_{i+1}}\cp_{i+1} +  \tfrac{t - t_{i+1}}{t_{i+2} - t_{i+1}}\cp_{i+2},
\end{eqnarray}
and
%\begin{eqnarray}
	$t_{i+1} = ||\cp_{i+1} - \cp_{i}||_2^\alpha + t_i$, $t_0 = 0$.
%\end{eqnarray}
Here, $\alpha$ ranges from 0 to 1. We choose $\alpha=0.5$ following~\cite{tan2014active}, which in theory produces splines without cusps or self-intersections~\cite{yuksel2011parameterization}.
%Let the control point sequence that defines a spline to be: $\cp_1,\cp_2,\cdots, \cp_n$, where $n$ is the total number of control points.
To make the spline a closed and $C^1$-continuous curve, we add three additional control points:\\[-6mm]
\begin{eqnarray}
	\cp_{N} &=& \cp_0\\
	\cp_{N+1} &=& \cp_0 + \tfrac{||\cp_{N-1} - \cp_0||_2}{||\cp_1 - \cp_0||_2} (\cp_1 - \cp_0)\\
	\cp_{-1} &=& \cp_0 + \tfrac{||\cp_{1} - \cp_0||_2}{||\cp_{N-1} - \cp_0||_2} (\cp_{N-1} - \cp_0). \qquad
\end{eqnarray}
%Thus, the final spline is parametrized by $\cp_0, \cp_1, \cdots, \cp_{n+2}$, with Eq.~\ref{eq:crs-spline}.

\subsection{Training}
\label{sec:training}
%For both poly-GCN and spline-GCN, the training data does not have ground truth control points sequence, we design the three(two) novel loss functions to optimize the network.
We train our model with two different loss functions. First, we train with a Point Matching Loss which we introduce in Subsec.~\ref{sec:matching}, and then fine-tune it with a Differentiable Accuracy Loss described in Subsec.~\ref{sec:diffrender}. Details and ablations are provided in Experiments.
\vspace{-3mm}
\subsubsection{Point Matching Loss}
\label{sec:matching}
\vspace{-1mm}
Typical point-set matching losses, such as the Chamfer Loss, assumed unordered sets of points (\ie they are permutation invariant). A polygon/spline, however, has a well defined ordering, which an ideal point set matching loss would obey. Assumuing equal sized and similarly ordered (clockwise or counter-clockwise) prediction and ground truth point sets, denoted as $\bp=\{p_0,p_1,\cdots,p_{K-1}\}$, and $\bp'=\{p'_0,p'_1,\cdots,p'_{K-1}\}$ respectively ($K$ is the number of points), we define our matching loss as:\\[-5mm]
\begin{eqnarray}
	L_{\mathrm{match}}(\bp, \bp') = \min_{j \in [0\cdots, K-1]}\sum_{i=0}^{K-1}{\lVert p_i - p'_{(j+i)\%K}\rVert}_1
\end{eqnarray}
Notice that this loss explicitly ensures an order in the vertices in the loss computation. Training with an unordered point set loss function, while maintaining the topology of the polygon could result in self-intersections, while the ordered loss function discourages it.
\vspace{-3.5mm}
\paragraph{Sampling equal sized point sets.} Since annotations may vary in the number of vertices, while our model always assumes $N$, we sample additional points along boundaries of both ground-truth polygons and our predictions. For Polygon-GCN, we uniformly sample $K$ points along edges of the predicted polygons, while for Spline-GCN, we sample $K$ points along the spline by uniformly ranging $t$ from $t_i$ to $t_{i+1}$. We also uniformly sample the same number of points along the edges of the ground-truth polygon. We use $K=1280$ in our experiments. Sampling more points would have a higher computational cost, while sampling fewer points would make curve approximation less accurate. Note that the sampling only involves interpolating the control points, ensuring differentiability.%We order the point set for both the ground-truth and prediction to be clockwise.

\vspace{-3mm}
\subsubsection{Differentiable Accuracy Loss}
\label{sec:diffrender}
%\begin{figure}[t!]
%\vspace{-2mm}
%\includegraphics[width=0.9\linewidth]{figs/diffrender_forwrd.jpg}
%\vspace{-7mm}
%\caption{We decompose polygon ABCDE into 3 triangle fans ABC, ACD and ADE and render them separately. We assign positive value for clock wise triangles(ABC, ACD) and negative value for counter clock wise triangles(ADE). Finally we sum over all the renderings. Only polygon area(green) will be filled and the outside polygon area(ADE) will be canceled. For more details, please refer to \cite{OpenGLredbook009}}
%\label{fig:diffrender_forward}
%\vspace{-2mm}
%\end{figure}
\begin{figure}[b]
\vspace{-3mm}
	%\vspace{-1cm}
	\begin{minipage}{0.41\linewidth}
	\vspace{-3mm}
	\begin{center}
		\includegraphics[width=1\textwidth]{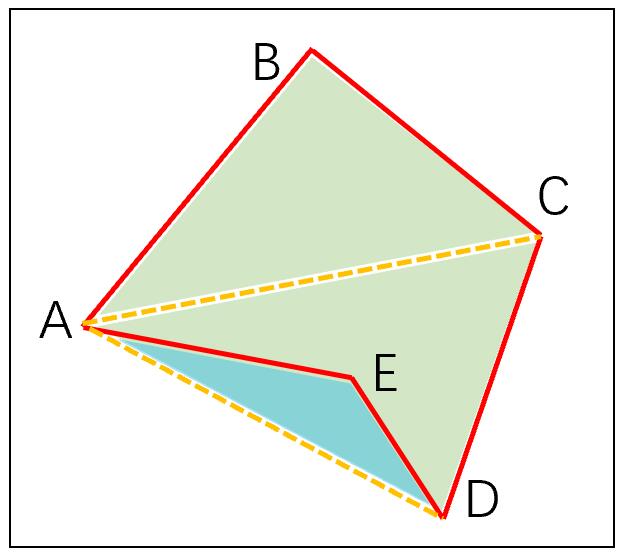}
	\end{center}
	\end{minipage}
	\hspace{3.5mm}
	%\vspace{-0.5cm}
	\begin{minipage}{0.5\linewidth}
\caption{\footnotesize We decompose the polygon ABCDE into 3 triangle fans ABC, ACD and ADE, and render them separately. We assign positive value for clock wise triangles (ABC, ACD) and negative value for the others (ADE). Finally we sum over all the renderings. The sum retains only the interior of the polygon.}
%	\vspace{-0.5cm}
	\label{fig:diffrender_forward}
	\end{minipage}
	\vspace{-4mm}
\end{figure}

Note that training with the point matching loss results in overly smooth predictions. To perfectly align the predicted polygon and the ground-truth silhouette, we employ a differentiable rendering loss, which encourages masks rendered from the predicted control points to agree with ground-truth masks by directly optimizing for accuracy. This has been used previously to optimize 3D mesh vertices to render correctly onto a 2D image~\cite{kato2018renderer,opendr2014}.
%While Loper et al~\cite{} and Kato et al~\cite{} exploit the differentiable render in 3D case to refine mesh, we are the first to introduce it into 2D case and demonstrate its effectiveness in improving segmentation performance.

The rendering process can be described as a function $R$;
%\begin{eqnarray}
	$M(\theta) = R(\bp(\theta))$,
%\end{eqnarray}
where $\bp$ is the sampled point sequence on the curve, and $M$ is the corresponding mask rendered from $\bp$. The predicted and the ground-truth masks can be compared by computing their difference with the $L1$ loss:
\begin{eqnarray}
	L_{\mathrm{render}}(\theta) = {\lVert M(\theta)  - M_{\mathrm{gt}}\rVert}_1
\end{eqnarray}
Note that $L_{\mathrm{render}}$ is exactly the pixel-wise accuracy of the predicted mask $M(\theta)$ with respect to the ground truth $M_{gt}$. We now describe the method for obtaining $M$ in the forward pass and back-propagating the gradients through the rendering process $R$, from $\frac{\partial L}{\partial M}$ to $\frac{\partial L}{\partial \bp}$ in the backward pass.

\vspace{-3mm}
\paragraph{Forward Pass:} We render $\bp$ into a mask using OpenGL. As shown in Fig.~\ref{fig:diffrender_forward}, we decompose the shape into triangle fans $\mathbf{f}_j$ and assign positive or negative values to their area based on their orientation. We render each face with the assigned value, and sum over the rendering of all the triangles to get the final mask. We note that this works for both convex and concave polygons~\cite{OpenGLredbook009}.

%Specifically, considering $\cp$ as a set of triangle faces $\textbf{f}_1$, $\textbf{f}_2$, ..., $\textbf{f}_n$, we draw each face with their direction and sum over all the triangles to get the final mask image.
%\begin{eqnarray}
%	\textbf{I}_i &=& sgn(\textbf{f}_i)R(\textbf{f}_i), \forall i=1,2,\cdots,N\\
%	\textbf{I} &=& \sum{\textbf{I}_i}.
%\end{eqnarray}

\vspace{-3mm}
\paragraph{Backward Pass:} The rendering process is non-differentiable in OpenGL due to rasterization, which truncates all float values to integers. However, following~\cite{opendr2014}, we compute its gradient with first order Taylor expansion. % As in the forward propagation, we have decomposed $\textbf{I}$ into a set of triangles $\textbf{f}_j$, thus, it is sufficient to analyze the gradients of $\textbf{I}_i$ on the triangle $\textbf{f}_i$. To be specific,
We reutilize the triangle fans from the decomposition in the forward pass (see Fig.~\ref{fig:diffrender_forward}) and analyze each triangle fan separately. Taking a small shift of the fan $\mathbf{f}_j$, we calculate the gradient w.r.t. the $j$-th triangle as:
\begin{eqnarray}
	\frac{\partial M_j}{\partial \mathbf{f}_j} = \frac{R(\mathbf{f}_j + \Delta t) - R(\mathbf{f}_j)}{\Delta t},
\end{eqnarray}
where $M_j$ is the mask corresponding to the fan $\mathbf f_j$. Here, $\Delta t$ can be either in the $x$ or $y$ direction. For simplicity, we let $\Delta t$ to be a $1$ pixel shift, which alleviates the need to render twice, and lets us calculate gradients by subtracting neighboring pixels. %Since the face $\textbf{f}_i$ is composed by 3 control points.
Next, we pass the gradient $\frac{\partial M_j}{\partial \mathbf{f}_j}$ to its three vertices $\mathbf{f}_{j,0}$, $\mathbf{f}_{j,1}$ and $\mathbf{f}_{j,2}$:
\vspace{-2mm}
\begin{eqnarray}
	\frac{\partial M_j}{\partial \mathbf{f}_{j,k}} = \sum_i{w_k^i \frac{\partial M_j^i}{\partial \mathbf{f}_j}}\quad k\in[0, 1, 2]
\end{eqnarray}
\vspace{-1mm}
where we sum over all pixels $i$. For the $i$-th pixel $M_j^i$ in the rendered image $M_j$, we compute its weight $w_0^i$, $w_1^i$ and $w_2^i$ with respect to the vertices of the face $\mathbf{f}_j$ as its barycentric coordinates. For more details, please refer to~\cite{opendr2014}.

\begin{figure}[t!]
\vspace{-3mm}
\includegraphics[width=\linewidth,trim=0 0 0 0,clip]{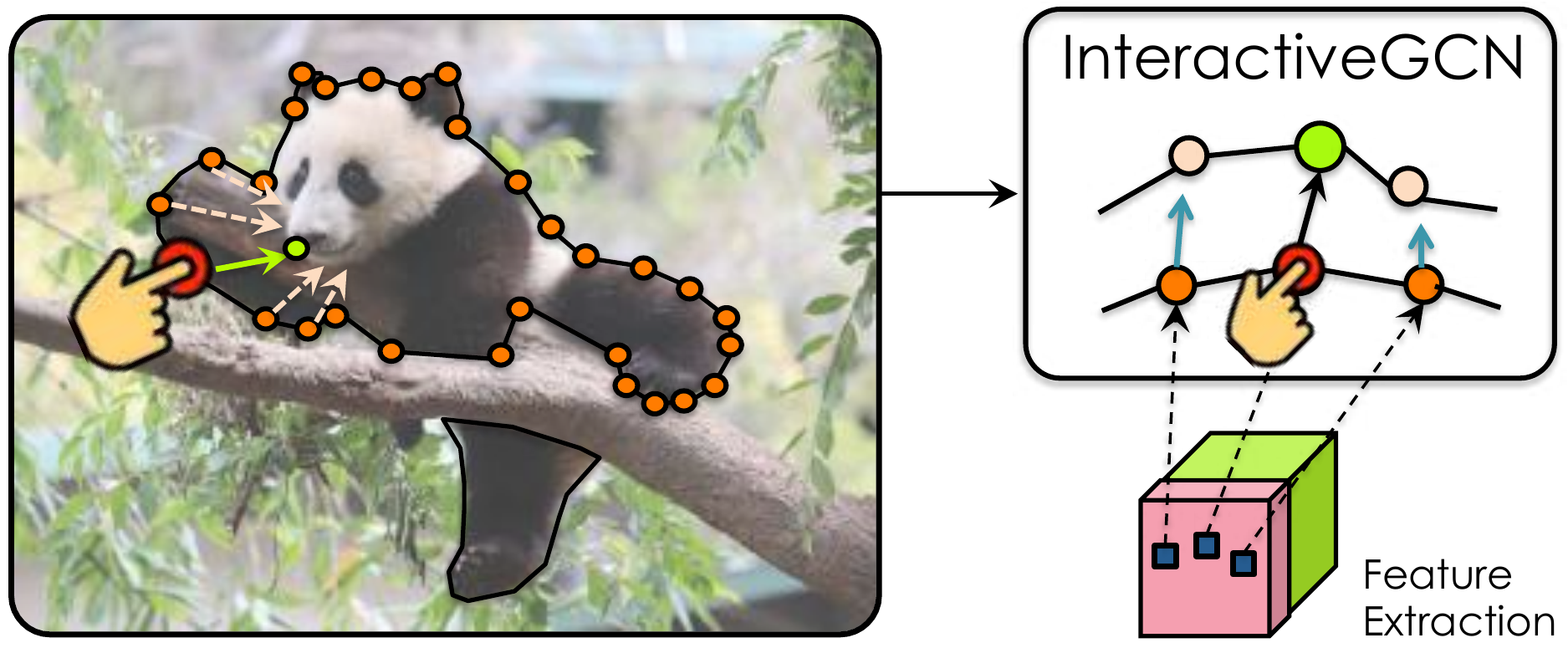}
\vspace{-7mm}
\caption{\footnotesize {\bf Human-in-the-Loop}: An annotator can choose any wrong control point and move it onto the boundary. Only its immediate neighbors ($k=2$ in our experiments) will be re-predicted based on this interaction.}
\label{fig:interactivegcn}
\vspace{-3mm}
\end{figure}

\subsection{Annotator in The Loop}
\label{sec:interactive}

The drawback of Polygon-RNN is that once the annotator corrects one point, \emph{all} of the subsequent points  will be affected due to the model's recurrent structure. This is often undesirable, as the changes can be drastic. In our work, we want the flexibility to change \emph{any} point, and further
%further be able to control \emph{how many} of the neighboring points can the change affect.
constrain that only the neighboring points can  change. As in Polygon-RNN, the correction is assumed to be in the form of drag-and-drop of a point.

To make our model interactive, we train another GCN that consumes the annotator's correction and predicts the relative shifts of the other control points. We refer to it as the \emph{InteractiveGCN}. We keep the network's architecture the same as the original GCN, except that we now append two additional dimensions to the corrected node's (say node $i$) input feature, representing the annotator's correction:
\begin{eqnarray}
	f_{i}^{0} = \mathrm{concat}\{F(x_i,y_i) ,x_i, y_i, \Delta x_i, \Delta y_i\},
\end{eqnarray}
where $(\Delta x_i, \Delta y_i)$ is the shift given by the annotator. For all other nodes, we set $(\Delta x_i, \Delta y_i)$ to zero. We do not perform iterative inference here.
Our InteractiveGCN allows a radius of influence by simply masking predictions of nodes outside the radius to 0. In particular, we let $k$ neighbors on either side of node $i$ to be predicted, \ie, $\cp_{(i-k)\%N}, \dots, \cp_{(i-1)\%N}, \cp_{(i+1)\%N}, \dots, \cp_{(i+k)\%N}$. We set $k=2$ in our experiments, while noting that in principle, the annotator could vary $k$ at test time. 

\iffalse
Our HumanGCN predicts a shift for each of the neighboring points, the remaining points' predictions are masked to $0$.
\begin{eqnarray}
	\cp_{j} = \cp_j + \mathbf{m}_j * \mathbf{p}_j, \forall j = 1, 2, \cdots, N,
\end{eqnarray}
where $\mathbf{p}_j$ is the predicted shift for the point $j$, and if annotator adjusts $i_{th}$ point, we set $\mathbf{m}_i = 1$, $\mathbf{m}_{(i-1)\%N} = 1$, $\mathbf{m}_{(i+1)\%N} = 1$, \dots, $\mathbf{m}_{(i-k)\%N} = 1$, $\mathbf{m}_{(i+ k)\%N} = 1$, and for other $\mathbf{m}_j$, we set them to zero. We set $k=2$ in our experiments, however, note that in principle the annotator can vary $k$ at test time. %\SF{what's this? what is p? is it correction?}
\fi

We train InteractiveGCN by mimicking an annotator that iteratively moves wrong control points onto their correct locations. %Specifically, we train the InteractiveGCN by iteratively feeding the (simulated) annotator's correction on InteractiveGCN's prediction as input, and the gradient back-propagates through this iterative procedure, helping the InteractiveGCN to incorporate possibly many user interactions.
We assume  that the annotator always chooses to correct the worst predicted point. This is computed by first aligning the predicted polygon with GT, by finding the minimum of our point matching loss (Sec.~\ref{sec:matching}). We then find the point with the largest manhattan distance to the corresponding GT point. The network is trained to move the neighboring points to their corresponding ground-truth positions. We then iterate between the annotator choosing the worst prediction, and training to correct its neighbors. In every iteration, the GCN first predicts the correction for the neighbors based on the last annotator's correction, and then the annotator corrects the next worst point.
We let the gradient back-propagate through the iterative procedure, helping the InteractiveGCN to learn to incorporate possibly many user interactions.
The training procedure is summarized in Alg.~\ref{algo:active}, where $c$ denotes the number of iterations. % and $\mathrm{Annotator}$ refers to the annotator moving the wrong control point from the prediction to its ground truth location.
%To train this network, we tried matching loss and differentiable render loss, the performance is almost the same, and we use matching loss at last due to its superior training speed.

\iffalse
\paragraph{Simulating the Annotator.}
The network corrects the neighboring points, taking annotator's correction as input.
After that, the simulated annotator choose the second points from the second predicted curve, and the network corrects the neighboring points again. This procedure runs iteratively until the IOU is greater than the threshold $T$ or the InteractiveGCN can't improve the prediction anymore. We consider the predicted splines achieving agreement above $T$ is satisfactory for the annotation. This is better than PolyRNN/PolyRNN++ as we allow annotator to stop correction at any time during the inference.
\fi

%\RestyleAlgo{boxruled}
%\SetAlFnt{\small}
\begin{algorithm}[t!]
%\SetAlgoLined
\begin{algorithmic}[1]
{\footnotesize
\While{not converged}
 \State (rawImage, gtCurve) = $\mathrm{Sample}$(Dataset)\;
 \State (predCurve, $F$) = $\mathrm{Predict}$(rawImage)\;
 \State data = []\;
 %\For{$i$ in range(c)}{
 \For{i in range(c)}
    \State corrPoint = $\mathrm{Annotator}$(predictedCurve)\;
  \State data += (predCurve, corrPoint, gtCurve, $F$)\;
  \State predCurve = $\mathrm{InteractiveGCN}$(predCurve, corrPoint)
  \State \Comment{Do not stop gradients}
  \EndFor
 \State $\mathrm{TrainInteractiveGCN}$(data)
 \EndWhile
}
 \end{algorithmic}
  \caption{\small{Learning to Incorporate Human-in-the-Loop}}
 \label{algo:active}
\end{algorithm}

%!TEX root = top.tex
% \vspace{-3mm}
\vspace{-2mm}
\section{Experimental Results}
\label{sec:results}
\vspace{-1mm}
% \vspace{-1mm}

\begin{table*}[t!]
\vspace{-3mm}
\begin{center}
{\footnotesize
\addtolength{\tabcolsep}{2.5pt}
\begin{tabular}{|l|c|c|c|c|c|c|c|c|c|c|}
\hline
Model & Bicycle & Bus & Person & Train & Truck & Motorcycle & Car & Rider & Mean \\
\hline 
\hline
Polygon-RNN++ &  57.38  &  75.99  &  68.45  & 59.65   &  76.31  &  58.26  & 75.68   & 65.65   &  67.17 \\
Polygon-RNN++ (with BS) &  63.06  &  81.38  &  72.41  & 64.28   &  78.90  &  62.01  & 79.08   & 69.95   &  71.38 \\
PSP-DeepLab &  67.18  &  83.81 &  72.62  & {\bf 68.76}   &  {\bf 80.48}  &  {\bf 65.94}  & 80.45   & 70.00   &  {\bf 73.66} \\
\hline
\hline
% Polygon-GCN-Master &  61.24  &  76.71   &  70.12  & 58.55   &  76.27  & 58.66   &  77.94    &  68.00   &  68.43 & < 2  \\
Polygon-GCN (MLoss) &63.68  & 81.42  &  72.25  & 61.45  &  79.88  & 60.86  & 79.84   &  70.17  &  71.19 \\
% \hspace{2mm}{+ Diff Render-Two Resnet} & \JUN{}  & \JUN{}  &  \JUN{}  & \JUN{}  &  \JUN{}  & \JUN{}  & \JUN{}   &  \JUN{}   &  \JUN{running} & < 2\\
\hspace{2mm}{+ DiffAcc} & 66.55 & 85.01 &  72.94  & 60.99  &  79.78  & 63.87 & 81.09   &  71.00   & 72.66 \\
\hline
Spline-GCN (MLoss) &  64.75  &  81.71   &  72.53  & 65.87  &  79.14  & 62.00  &  80.16   & 70.57   &  72.09  \\
% \hspace{2mm}{+ Coarse to Fine} &  63.16  &  80.00   &  71.94  & 57.10  &  78.02  & 60.43  &  79.44   &  69.93   &  70.00  & < 2  \\
% \hspace{2mm}{+ PSP instead Skip} &  63.49  &  79.37   &  71.67  & 60.69  &  79.51  & 61.17  &  79.38  &  69.94   &  70.65 & < 2  \\
% \hspace{2mm}{+ Boundary Prediction} &  64.75  &  81.71   &  72.53  & 65.87  &  79.14  & 62.00  &  80.16   & 70.57   &  72.09 & < 2   \\
% \hspace{2mm}{+ Diff Render-Two Resnet} &  65.91  & 82.69 &  73.38  & 66.79 &  79.72  & 63.09 &  80.72  & 71.64 &  72.99 & < 2  \\
\hspace{2mm}{+ DiffAcc} &  \textbf{67.36}  & \textbf{85.43} &  \textbf{73.72}  & 64.40  &  80.22  & 64.86  &  \textbf{81.88}  & \textbf{71.73} &  \textbf{73.70}\\
\hline
\end{tabular}
\vspace{-3mm}
\caption{\footnotesize {\bf Automatic Mode on Cityscapes.} We compare our Polygon and Spline-GCN to Polygon-RNN++ and PSP-DeepLab. Here, \emph{BS} indicates that the model uses beam search, which we do not employ.} %Note that PSP-DeepLab uses a more powerful backbone architecture than all other models (which use Resnet). }
\label{tbl:gt_boxes}
}
\end{center}
\end{table*}

\begin{table*}[t!]
\vspace{-4mm}
\vspace{-2mm}
\begin{minipage}{0.374\linewidth}
\begin{center}
{\footnotesize
\addtolength{\tabcolsep}{-2.0pt}
\begin{tabular}{|l|c|c|c|}
\hline
Model & mIOU & F at 1px & F at 2px  \\
\hline
Polyrnn++ (BS) & 71.38  & 46.57 & 62.26\\
\hline
PSP-DeepLab & 73.66  & 47.10 & 62.82 \\
% Spline-GCN-Two Resnet& \textbf{73.00}  & \textbf{47.53} & \textbf{63.16}\\
\hline
Spline-GCN& \textbf{73.70}  & \textbf{47.72} & \textbf{63.64}\\
\hline \hline
DEXTR & 79.40  & 55.38 & 69.84\\
% \hline
% Dextr-GCN-Two Resnet & \textbf{79.28}  & \textbf{56.59} & \textbf{70.70}\\
\hline
Spline-GCN-EXTR& \textbf{79.88}  & \textbf{57.56} & \textbf{71.89}\\
\hline
\end{tabular}
\vspace{-3mm}
\caption{\footnotesize{\bf Different Metrics}. We report IoU \& F boundary score. We favorably cross-validate PSP-DeepLab and DEXTR \emph{for each metric} on val. \emph{Spline-GCN-EXTR} uses extreme points as additional input as in DEXTR.}. 
\label{tbl:F-seg}
}
\end{center}
\end{minipage}
\hspace{1.7mm}
\begin{minipage}{.275\linewidth}
\vspace{-2mm}
\begin{center}
{\footnotesize
\addtolength{\tabcolsep}{-3pt}
\begin{tabular}{|l|c|c|}
\hline
Model & Spline & Polygon\\
\hline
GCN   & 68.55    &  67.79\\
\hline
\hspace{0.2mm}{+ Iterative Inference} &  70.00  &  70.78  \\
\hline
% \hspace{0.2mm}{+ PSP instead Skip}  &  70.65 & \JUN{TBA} \\
% \hline
\hspace{0.2mm}{+ Boundary Pred.} &  72.09 &  71.19  \\
\hline
\hspace{0.2mm}{+ DiffAcc} &  \textbf{73.70}& 72.66\\
\hline
\end{tabular}
\vspace{-2mm}
\caption{\footnotesize {\bf Ablation study} on Cityscapes. We use 3 steps when performing iterative inference. \emph{Boundary Pred} adds the boundary prediction branch to our CNN.}
\label{tbl:Ablation}
}
\end{center}
\end{minipage}
\hspace{1.7mm}
\begin{minipage}{0.305\linewidth}
\vspace{-6mm}
\begin{center}
{\footnotesize
\addtolength{\tabcolsep}{-1pt}
\begin{tabular}{|l|c|}
\hline
Model & Time(ms)\\
\hline
Polygon-RNN++ & 298.0\\
Polygon-RNN++ (Corr.) & 270.0\\
\hline
PSP-Deeplab & 71.3\\
\hline
Polygon-GCN & 28.7 \\
Spline-GCN  & 29.3\\
Polygon-GCN (Corr.) & 2.0\\
Spline-GCN (Corr.) & 2.6\\
\hline
\end{tabular}
\vspace{-2.8mm}
\caption{\footnotesize {\bf Avg. Inference Time} per object. We are $10\times$ faster than Polygon-RNN++ in forward pass, and $100\times$ for every human correction. }
\label{tbl:inference-time}
}
% {\footnotesize
% \addtolength{\tabcolsep}{2.5pt}
% \begin{tabular}{|l|c|c|}
% \hline
% N Points & Poly-GCN & Spline-GCN\\
% \hline
% 20 & 68.16 & \textbf{68.56}\\
% \hline
% 40 & 68.43 & \textbf{69.56}\\
% \hline
% \end{tabular}
% \vspace{-2mm}
% \caption{Comparison between different number of clicks.}
% \label{tbl:poly-spline}
% }
\end{center}
\end{minipage}
\vspace{-6mm}
\end{table*}

In this section, we extensively evaluate our Curve-GCN for both in-domain and cross-domain instance annotation. We use the Cityscapes dataset~\cite{cityscapes} as the main benchmark to train and test our model. We analyze both automatic and interactive regimes, and compare to state-of-the-art baselines for both. 
For cross-domain experiments, we evaluate the generalization capability of our Cityscapes-trained model on the KITTI dataset~\cite{kitti} and four out-of-domain datasets, ADE20K~\cite{ade20k}, Aerial Rooftop~\cite{rooftop}, Cardiac MR~\cite{cardiacmr}, and ssTEM~\cite{sstem}, following Polygon-RNN++~\cite{polyrnnpp}.

To indicate whether our model uses polygons or splines, we name them Polygon-GCN and Spline-GCN, respectively.

%For in-domain experiments, we provide results on Cityscapes~\cite{cityscapes} on both automatic and interactive instance segmentation in the Subsec.~\ref{sec:in-domain-annot}. We compare our results with state-of-arts boundary based segmentation method Polygon-RNN++~\cite{polyrnnpp} as well as state-of-arts pixel-wised based segmentation methods DeepLab and Dextr~\cite{dextr,ChenPKMY18}.  Ablation studies are also conducted to demonstrate the effectiveness of each component in our model. For cross-domain experiments, we evaluate the generalization capability of our model on the KITTI dataset~\cite{kitti} and four out-of-domain datasets, ADE20K~\cite{ade20k}, Aerial Rooftop~\cite{rooftop}, Cardiac MR~\cite{cardiacmr} and ssTEM~\cite{sstem} following Polygon-RNN++~\cite{polyrnnpp} in the Subsec~\ref{sec:cross-domain}.
 
  \vspace{-4mm}
\paragraph{Image Encoder:} Following Polygon-RNN++~\cite{polyrnnpp}, we use the ResNet-50 backbone architecture as our image encoder. %, unless otherwise stated.  %DeepLab's when comparing to DeepLab and DEXTR. 

 \vspace{-4mm}
\paragraph{Training Details:}
% \SF{provide details of the model's architecture. how many GCN, how many steps of inference?}
We first train our model via the matching loss, followed by fine-tuning with the differentiable accuracy loss. The former is significantly faster, but has less flexibility, \ie points are forced to exactly match the GT points along the boundary. Our differentiable accuracy loss provides a remedy as it directly optimizes for accuracy. However, since it requires a considerably higher training time we employ it only in the fine-tuning stage. %On cross-domain experiments, we take the model train on Cityscapes
For speed issues we use the matching loss to train the InteractiveGCN. We use a learning rate of 3e-5 which we decay every 7 epochs. 

We note that the Cityscapes dataset contains a significant number of occluded objects, which causes many objects to be split into disconnected components. Since the matching loss operates on single polygons, we train our model on single component instances first. We fine-tune with the differentiable accuracy loss on \emph{all} instances.

%We use a learning rate of 3e-5 which we decay for every 7 epochs. 
%We do inference iteratively three times.

\vspace{-4mm}
\paragraph{Baselines:} Since Curve-GCN operates in two different regimes, we compare it with the relevant baselines in each. For the automatic mode, we compare our approach to Polygon-RNN++~\cite{polyrnnpp}, and PSP-DeepLab~\cite{ChenPKMY18, pspnet}. We use the provided DeepLab-v2 model by~\cite{dextr}, which is pre-trained on ImageNet, and fine-tuned on PASCAL for semantic segmentation. We stack Pyramid scene parsing ~\cite{pspnet} to enhance performance. 
For the interactive mode, we benchmark against Polygon-RNN++ and DEXTR~\cite{dextr}. %We fine-tune both DeepLab and DEXTR on the Cityscapes dataset. 
%For fairness, when comparing with Polygon-RNN++ we use the ResNet-50 backbone architecture as our image encoder, and DeepLab's when comparing to DeepLab and DEXTR. 
We fine-tune both PSP-DeepLab and DEXTR on the Cityscapes dataset. We cross-validate their thresholds that decide between foreground/background on the validation set.

 \vspace{-4mm}
\paragraph{Evaluation Metrics:} We follow Polygon-RNN~\cite{polyrnn} to evaluate performance by computing Intersection-over-Union (IoU) of the predicted and ground-truth masks. However, as noted in~\cite{geodesic14}, IoU  focuses on the full region and is less sensitive to the inaccuracies along the object boundaries. We argue that for the purpose of object annotation boundaries are very important -- even slight deviations may not escape the eye of an annotator. We thus also compute the Boundary F score~\cite{Perazzi_CVPR_2016} which calculates precision/recall between the predicted and ground-truth boundary, by allowing some slack wrt misalignment.  Since Cityscapes is finely annotated, we report results at  stringent thresholds of 1 and 2 pixels.

% \vspace{-3mm}
%\paragraph{Baselines.}  

%For the automatic mode, where the model directly predicts boundary curve without human intervention, we measure the performance with two metircs: intersection over union (IOU) and Boundary F score, which is inherited from Davis Benchmark~\cite{Perazzi_CVPR_2016}. Specifically, we calculates precision/recall between the predicted boundary and the ground truth boundary, which is expanded using binary dilation with a pixel threshold. We report results where the threshold is 1 and 2.

%\paragraph{Robust Spline GCN: } \HUAN{ ;(.. to delete later if anything works.} There are also many polygon annotated dataset which don't have multiply components instances. To make \textbf{Mitosis-Spline-GCN} robust, and also exploit multiply components data, we do prediction iteratively. As shown in Fig.~\ref{}. Instead finetune \textbf{Master-Spline-GCN}, we freeze gradient and take output of \textbf{Master-Spline-GCN} as initial position. We train a second \textbf{Master-Spline-GCN} by Diffrender loss. As shown in visulization Fig.~\ref{}, the second spline GCN does local fine tunning.

\vspace{-1mm}
\subsection{In-Domain Annotation}
\label{sec:in-domain-annot}
We first evaluate our model when both training and inference are performed on Cityscapes~\cite{cityscapes}. This dataset contains 2975/500/1525 images for training, validation and test, respectively. For a fair comparison, we follow the same split and data preprocessing procedure as in Polygon-RNN++~\cite{polyrnn}.

%\paragraph{Cityscapes:}To deal with occlusions, each ground truth instance can have multiple components. There are XXX instances which have only single components and XXX instances with multiply components. In total, there are XX instances and YY components, which leads to ZZ components per data. As in Polygon-RNN ~\cite{polyrnn} and Polygon-RNN++~\cite{polyrnnpp}, authors made segmentation task harder by cropping bounding box as square rather than stretching the bounding box rectangle so that each bounding box might contains multiply instances. 
 
% \paragraph{Baselines:} We compare our model with PolyRNN++~\cite{polyrnnpp}, which is state-of-the-art baseline for boundary-based instance segmentation. For the automatic mode, we additionally compare with state-of-the-art pixel-based baseline, DeepLab~\cite{ChenPKMY18}. We also compare with DEXTR~\cite{dextr}, which utilizes four clicks of extreme points from human.

\begin{figure*}[t!]
\vspace{-2mm}
\addtolength{\tabcolsep}{-3.6pt}
\begin{tabular}{ccc}
\includegraphics[height=2.18cm,trim=0 0 0 0,clip]{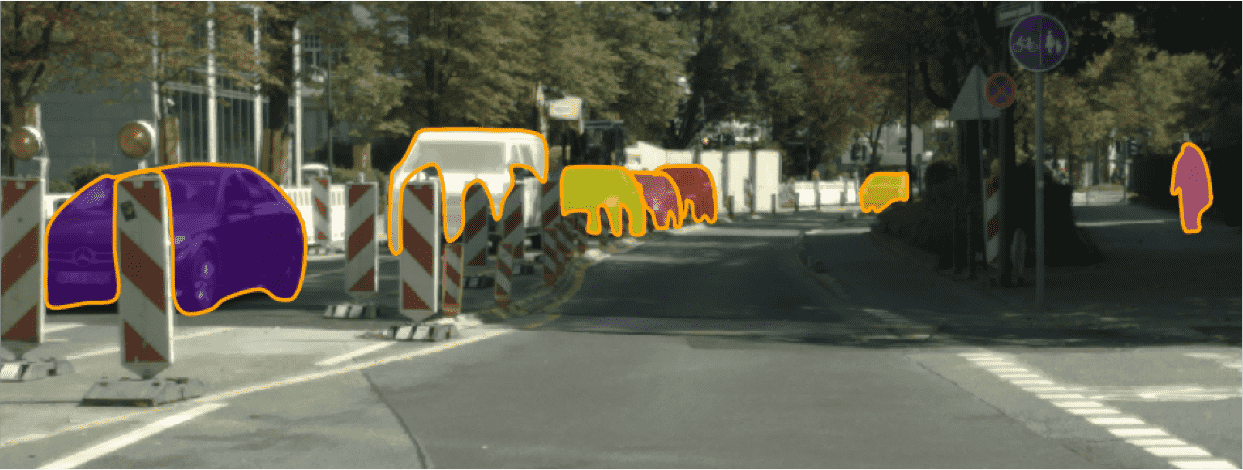} &
\includegraphics[height=2.18cm,trim=0 0 0 0,clip]{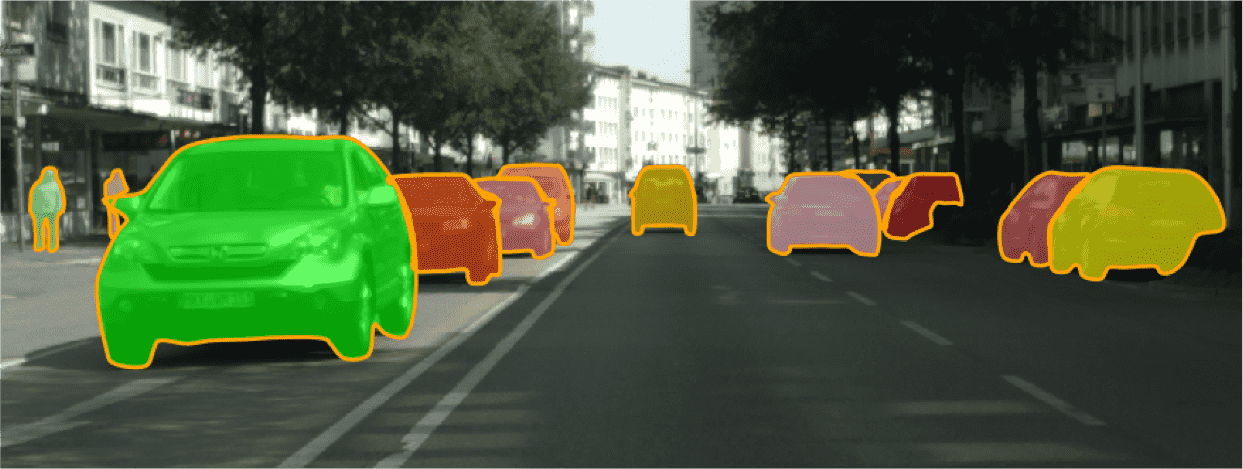} &
\includegraphics[height=2.18cm,trim=180 180 142 125,clip]{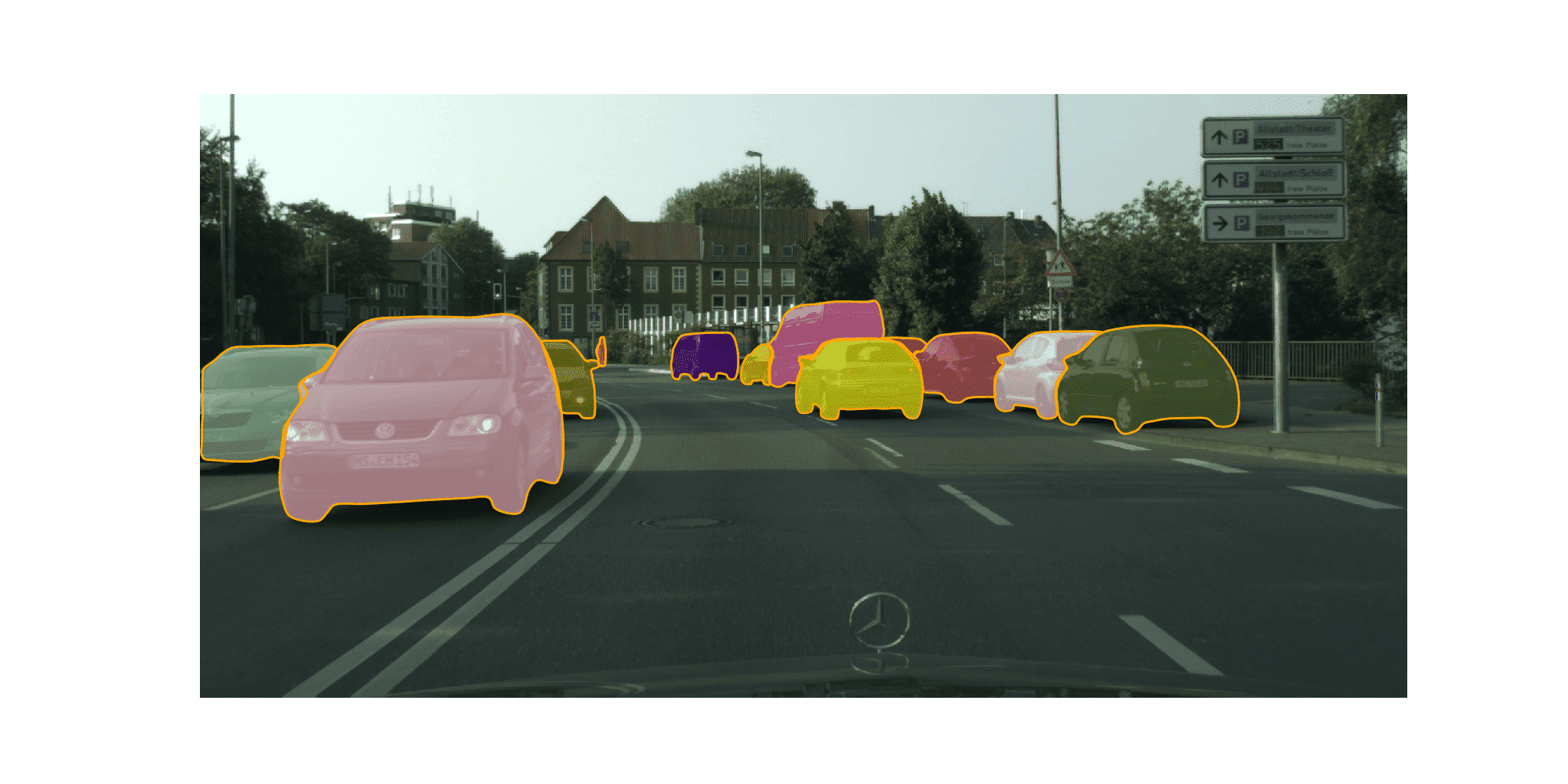} 
\end{tabular}
\vspace{-5mm}
\caption{\footnotesize{\bf Automatic Mode on Cityscapes}. The input to our model are bounding boxes for objects. }
\label{fig:quantitive_results1}
\vspace{-2mm}
\end{figure*}

\begin{figure*}[t!]
\addtolength{\tabcolsep}{-3.2pt}
\begin{tabular}{ccccccc}
\includegraphics[height=2cm,trim=110 90 80 100,clip]{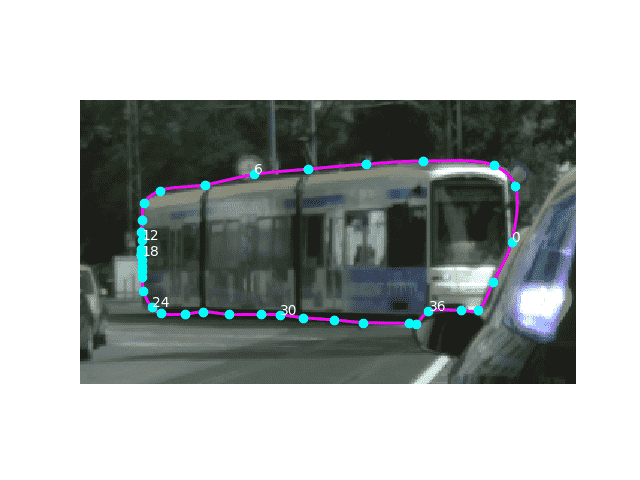} &
\includegraphics[height=2cm,trim=100 90 80 90,clip]{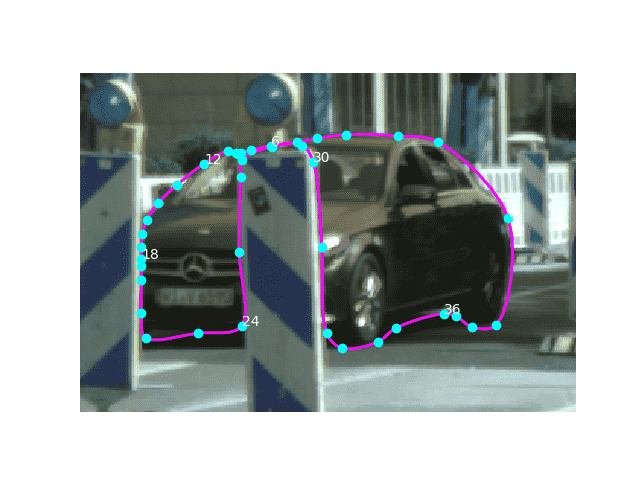} &
\includegraphics[height=2cm,trim=100 90 80 90,clip]{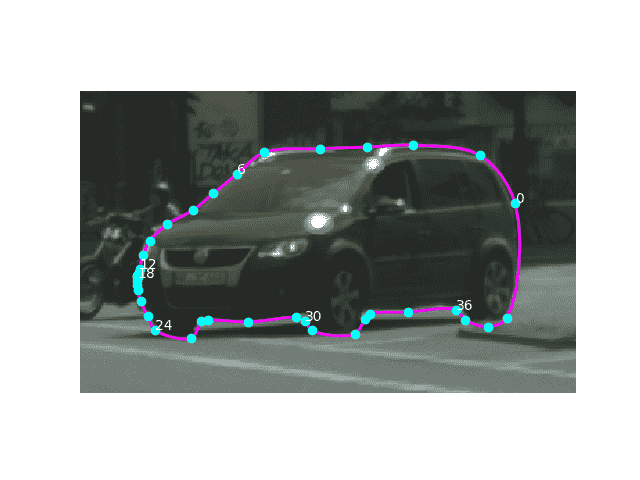} &
\includegraphics[height=2cm,trim=230 70 210 70,clip]{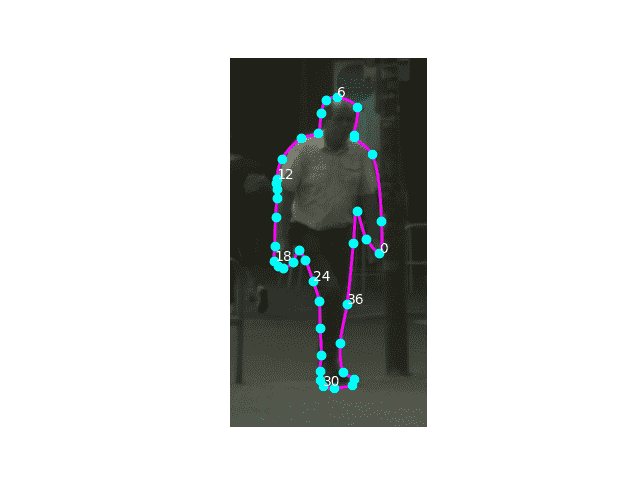} &
\includegraphics[height=2cm,trim=140 70 120 70,clip]{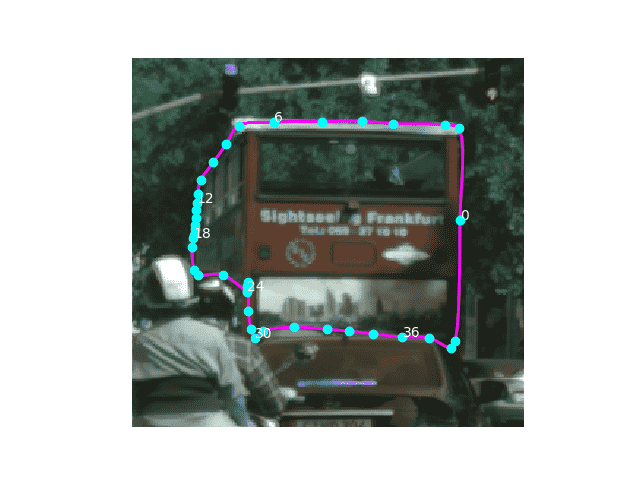} &
\includegraphics[height=2cm,trim=200 90 180 70,clip]{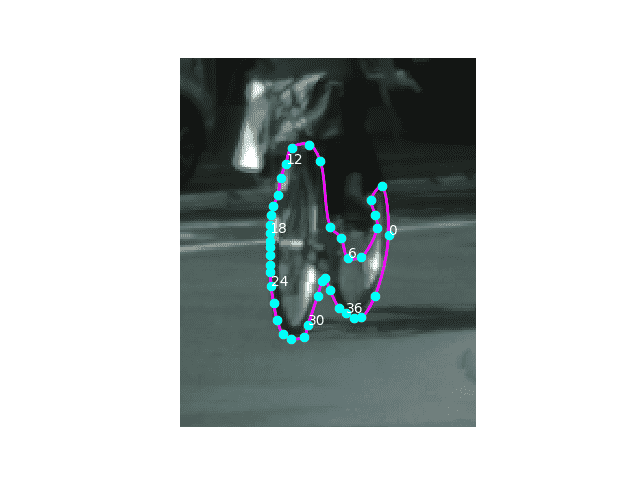} &
\includegraphics[height=2cm,trim=150 90 130 70,clip]{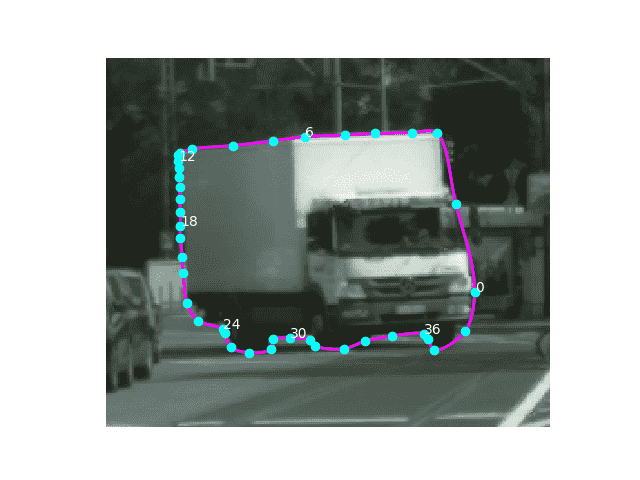} \\
\includegraphics[height=2cm,trim=110 90 80 100,clip]{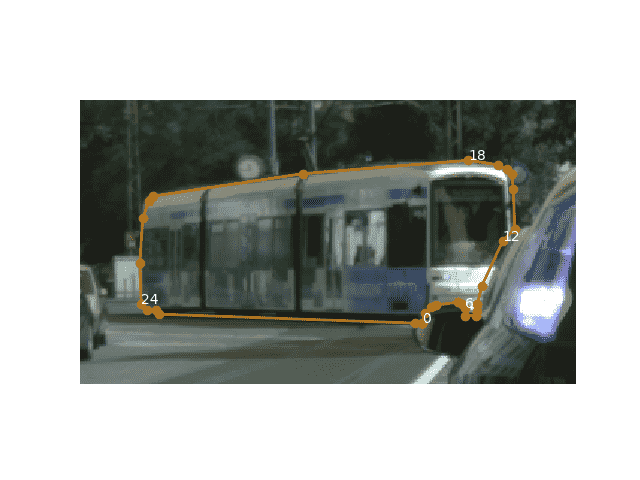} &
\includegraphics[height=2cm,trim=100 90 80 90,clip]{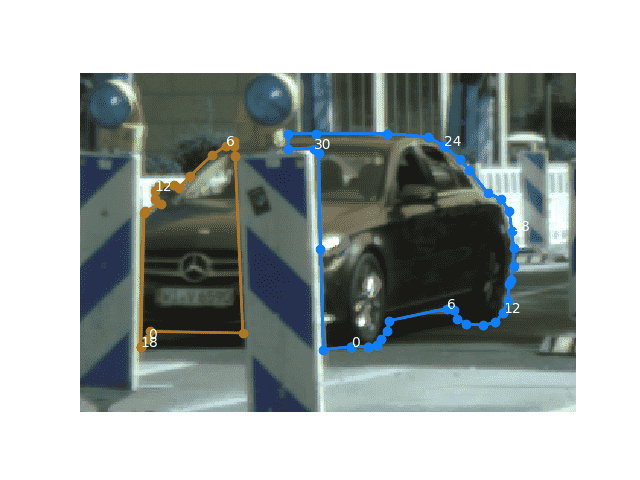} &
\includegraphics[height=2cm,trim=100 90 80 90,clip]{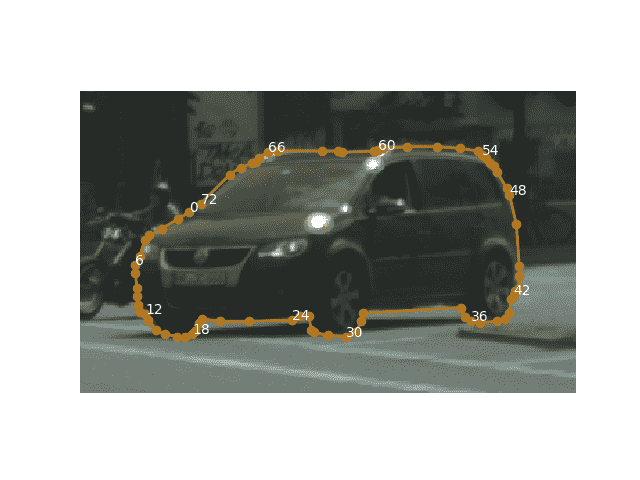} &
\includegraphics[height=2cm,trim=230 70 210 70,clip]{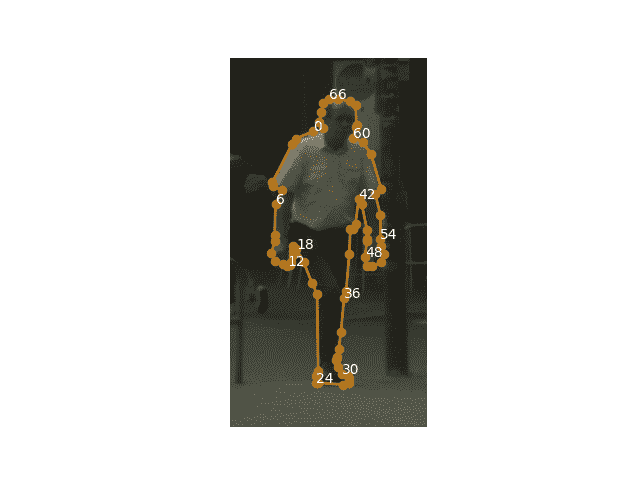} &
\includegraphics[height=2cm,trim=140 70 120 70,clip]{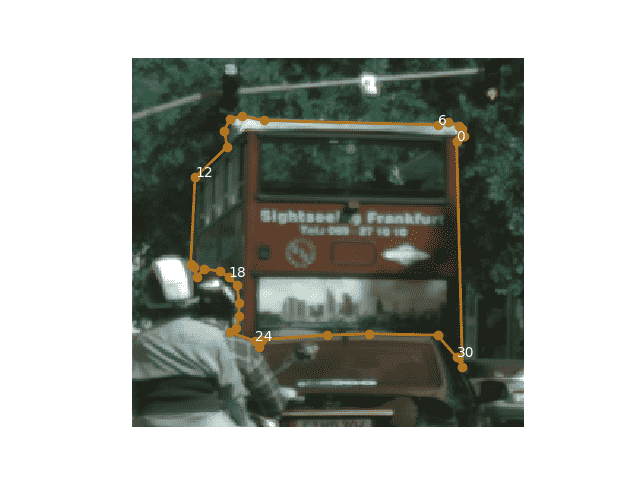} &
\includegraphics[height=2cm,trim=200 90 180 70,clip]{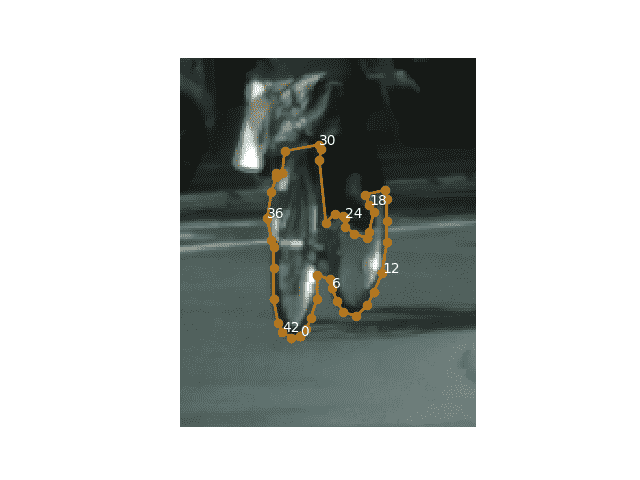} &
\includegraphics[height=2cm,trim=150 90 130 70,clip]{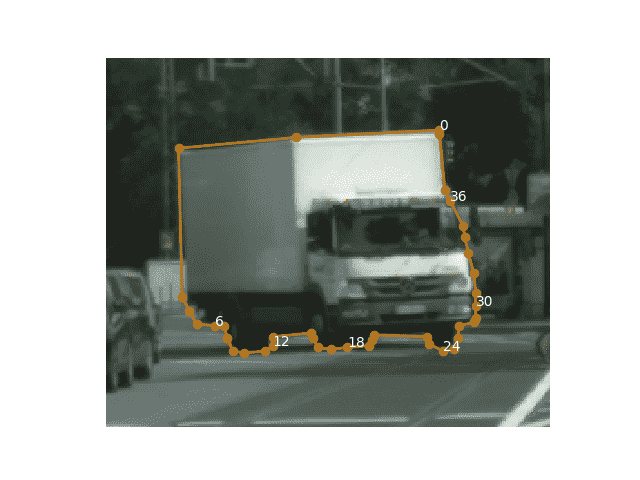}
\end{tabular}
\vspace{-5mm}
\caption{\footnotesize {\bf Automatic mode on Cityscapes}. We show results for individual instances. {\bf (top)} Spline-GCN, {\bf (bottom)} ground-truth. We can observe that our model fits object boundaries accurately, and surprisingly finds a way to ``cheat" in order to annotate multi-component instances.}
\label{fig:instance_examples}
\end{figure*}

\begin{figure*}[t!]
\vspace{-2mm}
\addtolength{\tabcolsep}{-4.2pt}
\begin{tabular}{ccccp{1mm}cccc}
\includegraphics[height=2.18cm,trim=0 0 0 0,clip]{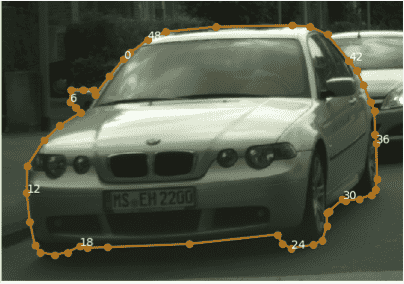} &
\includegraphics[height=2.18cm,trim=0 0 0 0,clip]{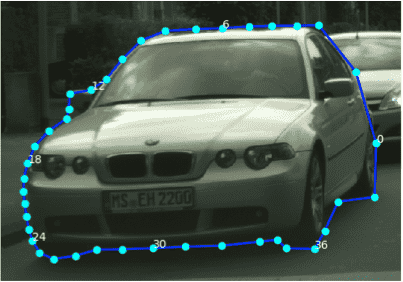} &
\includegraphics[height=2.18cm,trim=0 0 0 0,clip]{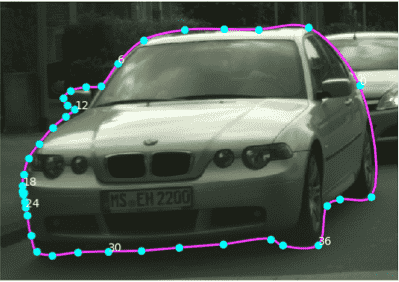} &
\includegraphics[height=2.18cm,trim=0 0 0 0,clip]{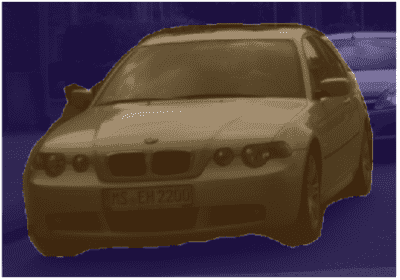} &
&
\includegraphics[height=2.18cm,trim=0 0 0 0,clip]{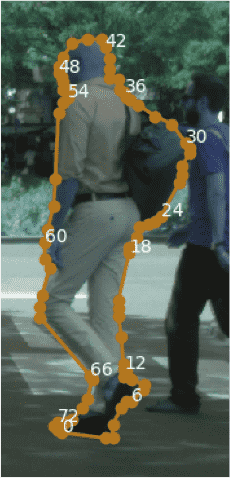} &
\includegraphics[height=2.18cm,trim=0 0 0 0,clip]{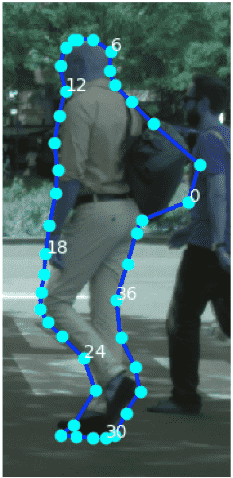} &
\includegraphics[height=2.18cm,trim=0 0 0 0,clip]{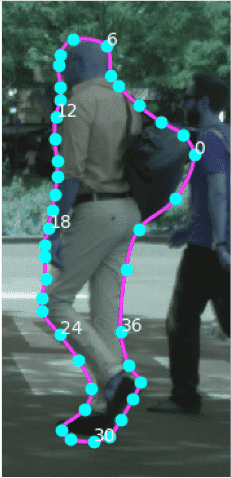} &
\includegraphics[height=2.18cm,trim=0 0 0 0,clip]{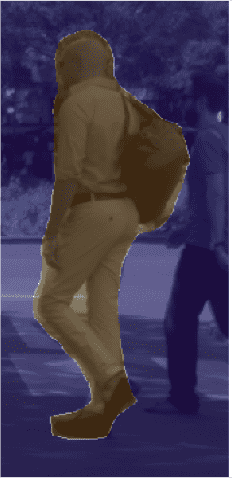} 
\end{tabular}
\vspace{-5mm}
\caption{\footnotesize {\bf Comparison in Automatic Mode}. From left to right: ground-truth, Polygon-GCN, Spline-GCN, PSP-DeepLab.}
\label{fig:quantitive_results2}
\vspace{-4mm}
\end{figure*}

\vspace{-4mm}
\paragraph{Automatic Mode:} %We first compare our approach with Polygon-RNN++, which uses the same backbone architecture.  The results are provided in the Table~\ref{tbl:gt_boxes} 
Table~\ref{tbl:gt_boxes} reports results of our Curve-GCN and compares with baselines, in terms of IoU. Note that PSP-DeepLab uses a more powerful image encoder, which is pretrained on PASCAL for segmentation. Our Spline-GCN outperforms Polygon-RNN++ and is on par with PSP-DeepLab. It also wins over Polygon-GCN, likely because most Cityscapes objects are curved. The results also show the significance of our differentiable accuracy loss  (diffAcc) which leads to large improvements over the model trained with the matching loss alone (denoted with MLoss in Table). Our model mostly loses against PSP-DeepLab on the \emph{train} category, which we believe is due to the fact that trains in Cityscapes are often occluded and broken into multiple components. Since our approach predicts only a single connected component, it struggles in such cases. 

Table~\ref{tbl:F-seg} compares models with respect to F boundary metrics. We can observe that while Spline-GCN is on par with PSP-DeepLab under the IoU metric, it is significantly better in the more precise F score. This means that our model more accurately aligns with the object boundaries. We show qualitative results in Fig~\ref{fig:quantitive_results1},~\ref{fig:instance_examples}, and~\ref{fig:quantitive_results2}. 

\vspace{-5mm}
\paragraph{Ablation Study:}
We study each component of our model and provide results for both Polygon and Spline-GCN in Table~\ref{tbl:Ablation}. Performing iterative inference leads to a significant boost, and adding the boundary branch to our CNN further improves performance.

\vspace{-4mm}
\paragraph{Additional Human Input:} In DEXTR~\cite{dextr}, the authors proposed to use 4 extreme points on the object boundary as an effective information provided by the annotator. Compared to just a box, extreme points require 2 additional clicks. We compare to DEXTR in this regime, and follow their strategy in how this information is provided to the model. To be specific, points (in the form of a heat map) are stacked with the image, and passed to a CNN. To compare with DEXTR, we use DeepLab-v2 in this experiment, as they do. 
We refer to our models with such input by appending \emph{EXTR}. %Results are reported in Table~\ref{tbl:more_clicks}. 

We notice that the image crops used in Polygon-RNN, are obtained by extracting an image inside a \emph{square box} (and not the actual box provided by the annotator). However, due to significant occlusion in Cityscapes, doing so leads to ambiguities, since multiple objects can easily fall in the same box. By providing 4 extreme points, the annotator more accurately points to the target object. To verify how much accuracy is really due to the additional two clicks, we also test an instantiation of our model to which the four corners of the bounding box are provided as input. This is still a 2-click (box) interaction from the user, however, it reduces the ambiguity of  which object to annotate. We refer to this model by appending \emph{BOX}.

Since DEXTR labels pixels and thus more easily deals with multiple component instances, we propose another instantiation of our model which still exploits 4 clicks on average, yet collects these differently. Specifically, we request the annotator to provide a box around \emph{each component}, rather than just a single box around the full object. On average, this leads to 2.4 clicks per object. This model is referred to with \emph{MBOX}. To match the 4-click budget, our  annotator clicks on the worst predicted boundary point for each component, which leads to 3.6 clicks per object, on average. 

Table~\ref{tbl:more_clicks} shows that in the extreme point regime, our model is already better than DEXTR, whereas our alternative strategy is even better, yielding an $0.8\%$ improvement overall with fewer clicks on average.  Our method also significantly outperforms DEXTR in the boundary metrics (Table~\ref{tbl:F-seg}).

\begin{table*}[t!]
\vspace{-2mm}
\begin{center}
{\footnotesize
\addtolength{\tabcolsep}{1.7pt}
\begin{tabular}{|l|c|c|c|c|c|c|c|c|c|c|}
\hline
Model & Bicycle & Bus & Person & Train & Truck & Mcycle & Car & Rider & Mean & \# clicks \\
\hline\hline
% Dextr Spline-GCN  & 70.55 & 83.46 & 77.33 & 69.66 & 83.16 & 69.90 & 84.02 & 74.04 &76.51  & 4\\
% \hspace{2mm}{+ Coarse to Fine} & 71.13 & 84.81 & 78.61 & 69.48 & 84.23 & 70.21 & 84.52 & 75.34 & 77.29 & 4\\
% \hspace{2mm}{+ Boundary Prediction} & 72.68 & 85.47 & 79.24 & 70.62 & 86.26 & 71.14 & 85.52 & 75.90 & 78.35 & 4 \\
% \hspace{2mm}{+ Diff Render} & \textbf{75.09} & \textbf{87.40} & \textbf{79.88} & 72.78 & \textbf{86.76} & \textbf{73.93} & \textbf{86.13} & \textbf{77.12} & \textbf{79.88} & 4 \\
Spline-GCN-BOX  & 69.53 & 84.40 & 76.33 & 69.05 & 85.08 & 68.75 & 83.80 & 73.38 & 76.29 & 2 \\
\hline
\hline
PSP-DEXTR &  74.42 &  87.30  &  79.30  &  73.51   &  85.42  & 73.69  &  85.57    &  76.24  & 79.40  & 4\\
\hline
%Spline-GCN-EXTR  & 72.68 & 85.47 & 79.24 & 70.62 & 86.26 & 71.14 & 85.52 & 75.90 & 78.35 & 4 \\
%\hspace{2mm}{+ Diff Render-Two Resnet} & 74.29 & 87.11 & 79.78 & 71.92 & 85.70 & 72.54 & 85.80 & 76.72 & 79.28 & 4 \\
{Spline-GCN-EXTR} & \textbf{75.09} & 87.40 & \textbf{79.88} & 72.78 & \textbf{86.76} & 73.93 & \textbf{86.13} & \textbf{77.12} & 79.88 & 4 \\
% \hline\hline
% DeepLab-PSP ($\bigtriangledown$)&  67.19 &  83.81  &  72.62  &  68.76   &  80.49  & 65.95  &  80.45    &  70.00  & 73.66  & 2\\
\hline \hline
Spline-GCN-MBOX & 70.45 & 88.02 & 75.87 & 76.35 & 82.73 & 70.76 & 83.32 & 73.49 & 77.62 & 2.4\\
\hspace{4mm}{+ One click} & 73.28 & \textbf{89.18} & 78.45 & \textbf{79.89} & 85.02 & \textbf{74.33} & 85.15 & 76.22 & \textbf{80.19} & 3.6\\
\hline
\end{tabular}
\vspace{-3mm}
\caption{\footnotesize {\bf Additional Human Input.} We follow DEXTR~\cite{dextr} and provide a budget of 4 clicks to the models. Please see text for details.}
\label{tbl:more_clicks}
}
\end{center}
\end{table*}

\begin{figure*}[t!]
\vspace{-7mm}
		\centering
%\begin{minipage}{0.49\linewidth}
%		\includegraphics[width=\textwidth]{./figs/human_in_loop/cityscapes_deeplab_40_master.pdf}
%		\caption{Cityscapes Deeplab 40 control points}
%\end{minipage}
%\begin{minipage}{0.49\linewidth}
%		\includegraphics[width=\textwidth]{./figs/human_in_loop/cityscapes_deeplab_20_master.pdf}
%		\caption{Cityscapes Deeplab 20 control points}
%		\end{minipage}
\begin{minipage}{0.665\linewidth}
\begin{minipage}{0.48\linewidth}
		\includegraphics[height=3.7cm,trim=50 0 70 30,clip]{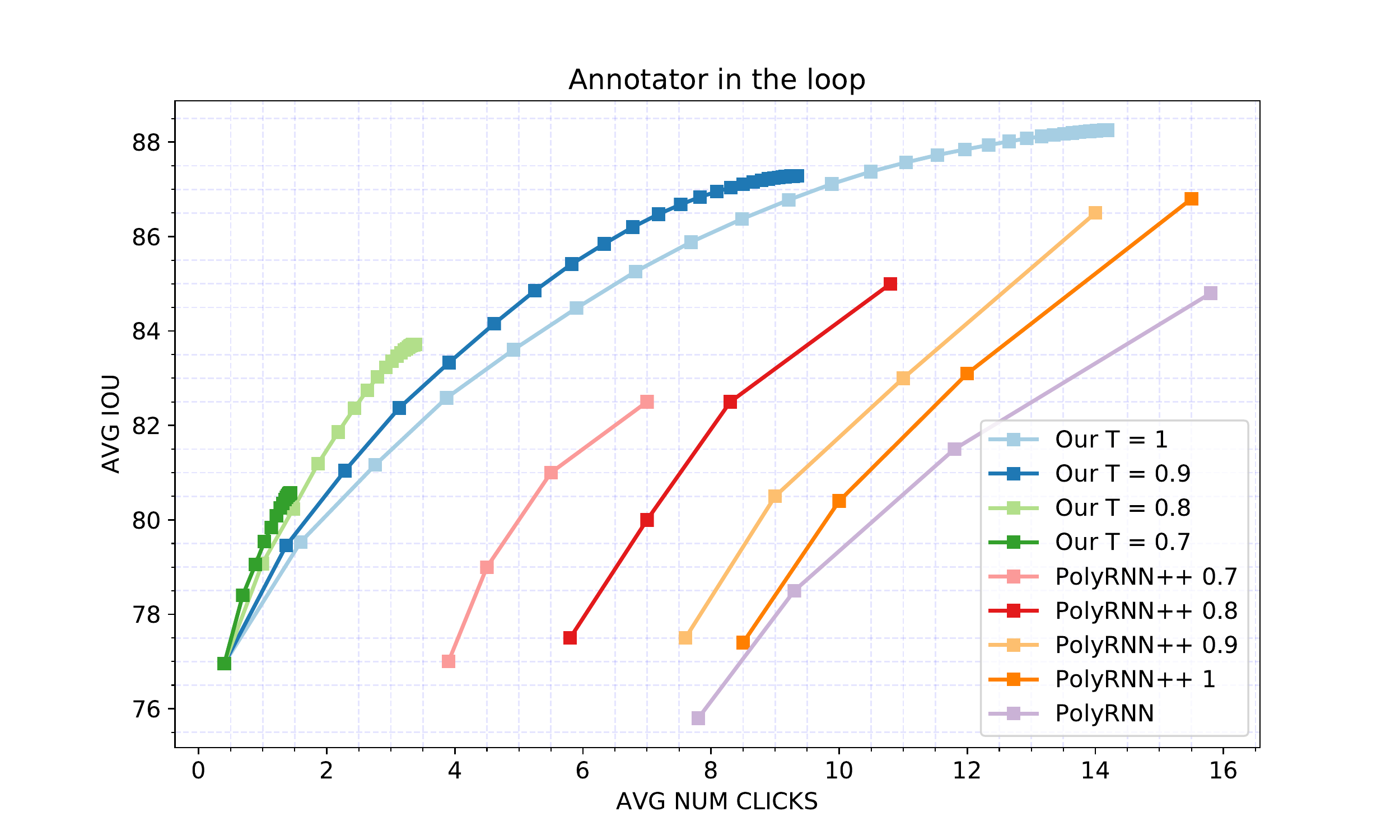}
		%\caption{Cityscapes Resnet 40 control points}
		\end{minipage}
\begin{minipage}{0.48\linewidth}
		\includegraphics[height=3.7cm,trim=40 0 70 30,clip]{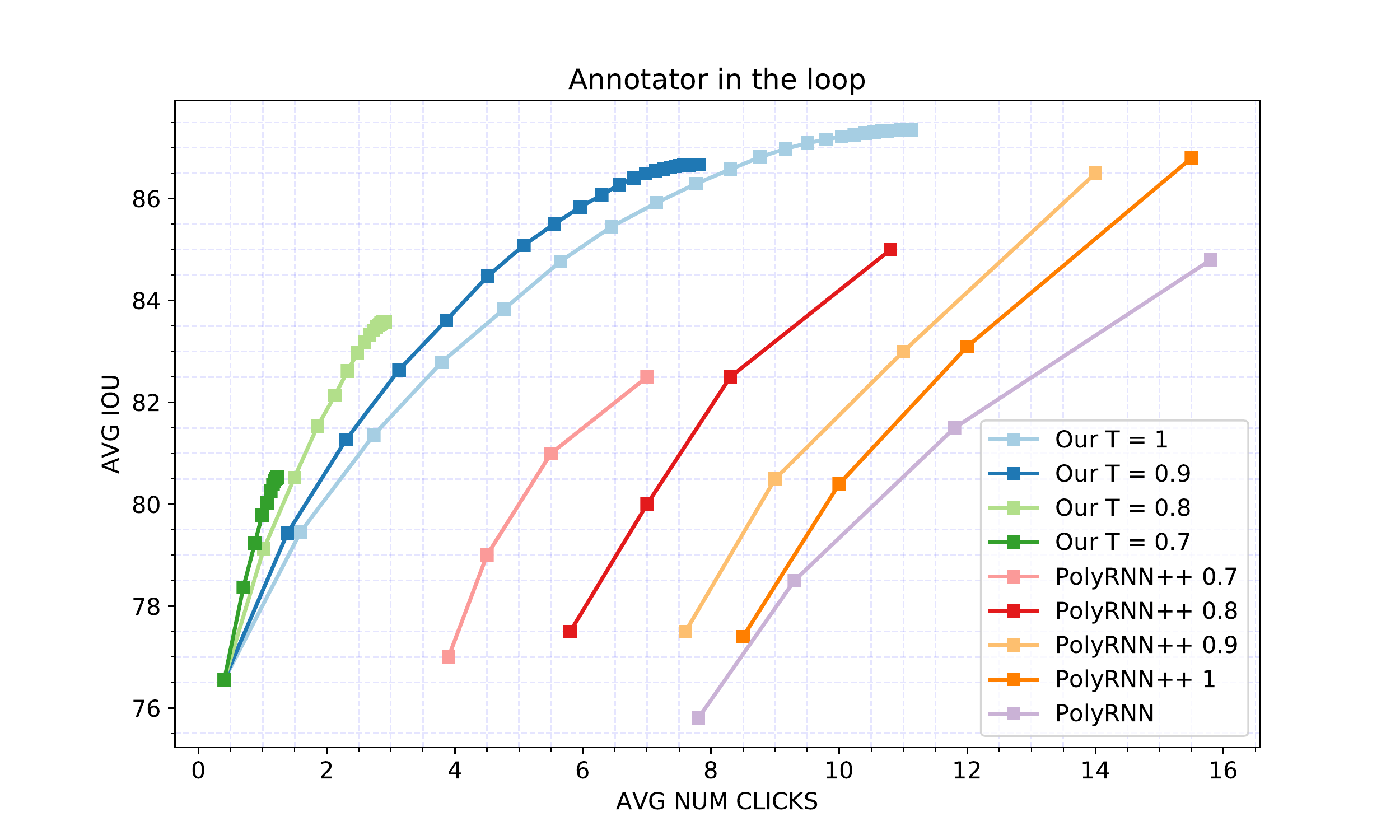}
		%\caption{Cityscapes Resnet 20 control points}
\end{minipage}
\vspace{-4mm}
	\caption{\footnotesize {\bf Interactive Mode on Cityscapes}: ({\bf left}) 40 control points, ({\bf right}) 20 control points.}
	\label{fig:human-in-loop}
\end{minipage}
\begin{minipage}{0.31\linewidth}
%\begin{minipage}{0.49\linewidth}
%\includegraphics[width=\textwidth,trim=50 0 50 0]{./figs/human_in_loop/kitti_resnet_40.pdf}
%\end{minipage}
\begin{minipage}{1\linewidth}
\includegraphics[height=3.7cm,trim=50 0 00 30,clip]{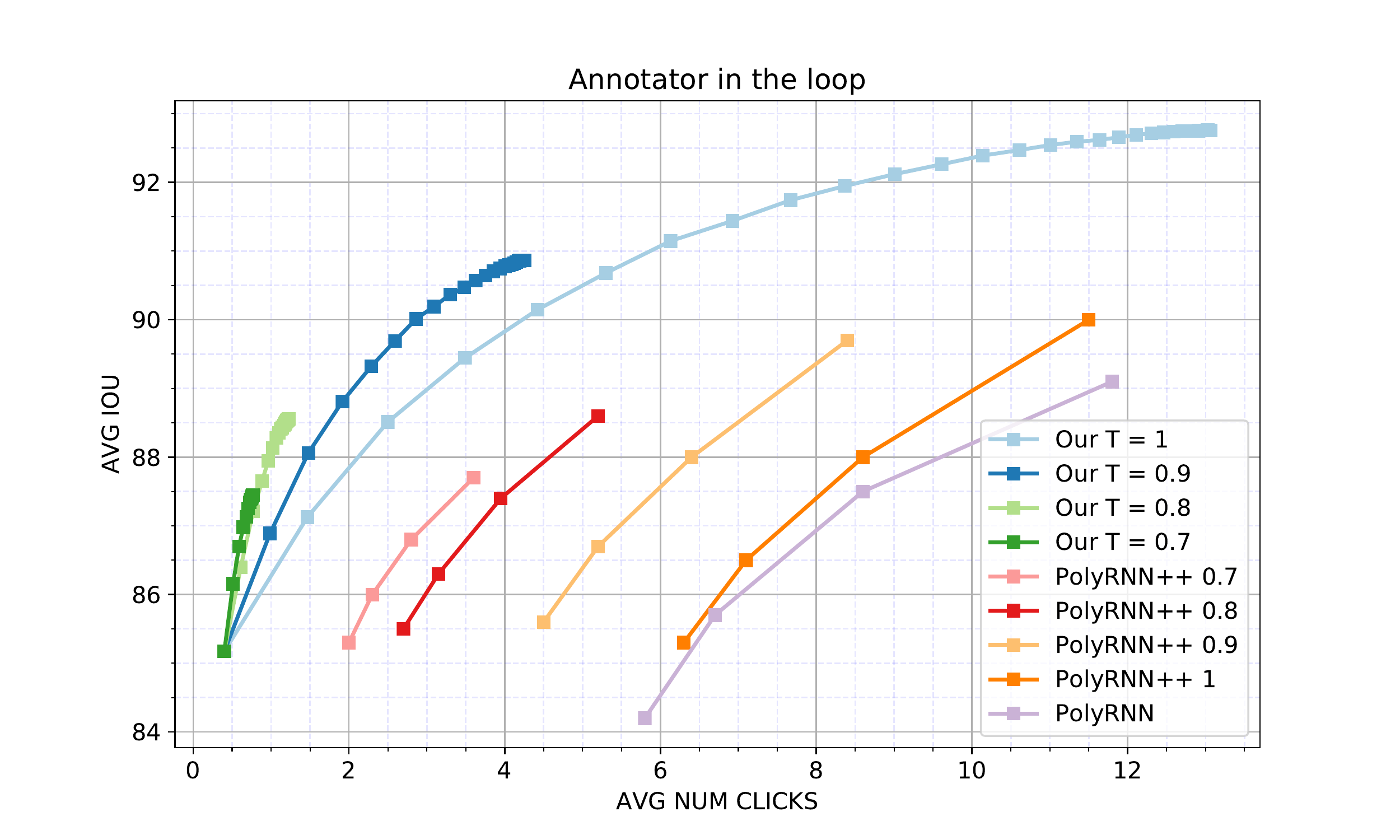}
\end{minipage}
\vspace{-5mm}
	\caption{\footnotesize{\bf Inter. Mode on KITTI}: 40 cps}
	\label{fig:human-in-loop-kitti}
\end{minipage}
\end{figure*}

\begin{figure*}[t!]
\vspace{-2.2mm}
\addtolength{\tabcolsep}{-4.95pt}
\begin{tabular}{cccp{1mm}cccccc}
\includegraphics[height=1.85cm,trim=180 95 145 105,clip]{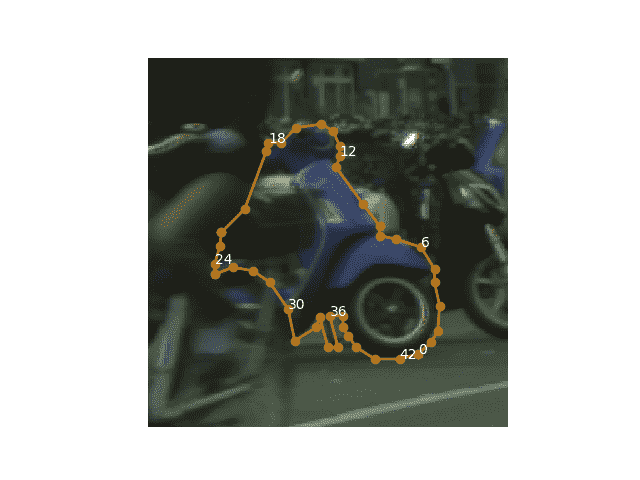} &
\includegraphics[height=1.85cm,trim=180 95 145 105,clip]{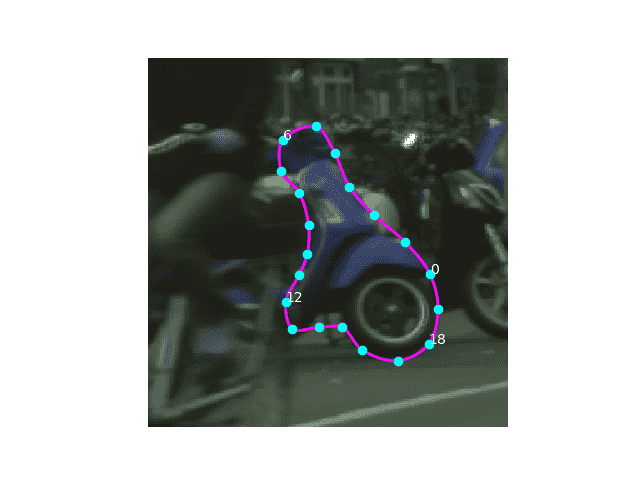} &
\includegraphics[height=1.85cm,trim=180 95 145 105,clip]{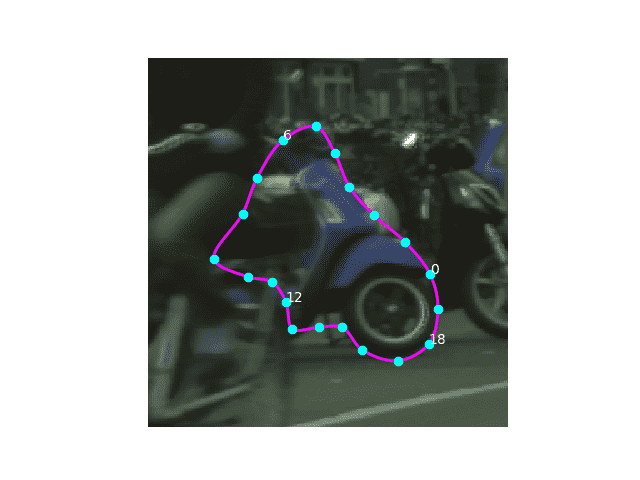} & &
\includegraphics[height=1.85cm,trim=100 55 92 56, clip]{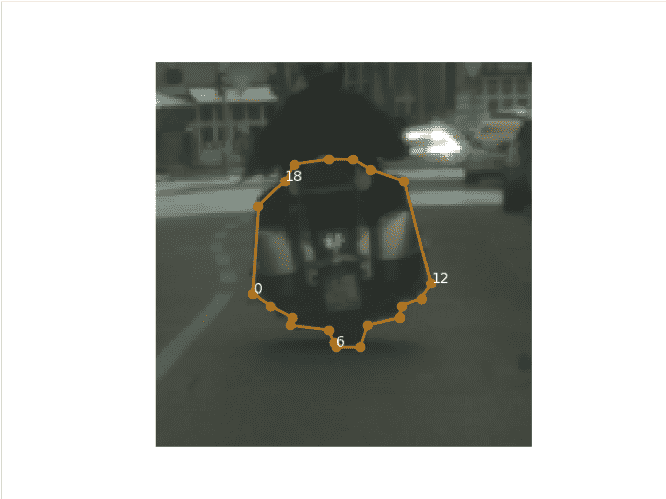} &
\includegraphics[height=1.85cm,trim=200 95 185 110,clip]{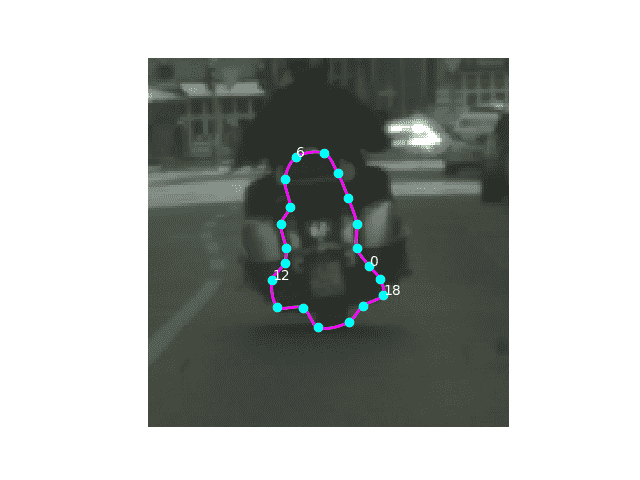} &
\includegraphics[height=1.85cm,trim=200 95 185 110,clip]{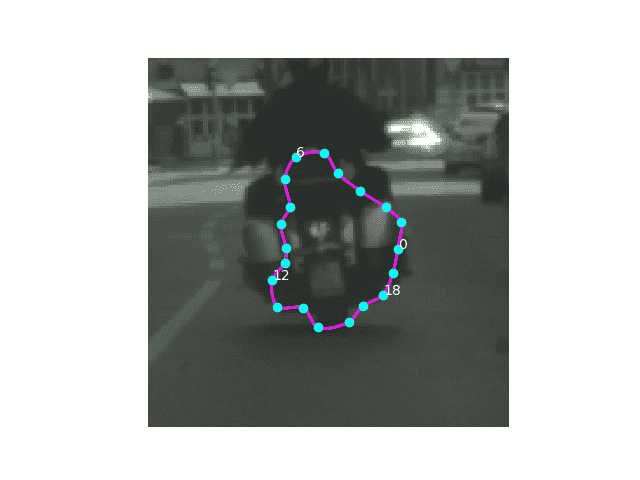} &
\includegraphics[height=1.85cm,trim=200 95 185 110,clip]{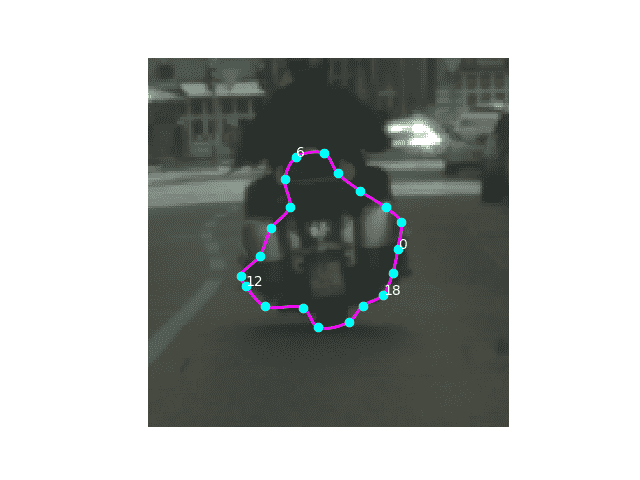} &
\includegraphics[height=1.85cm,trim=200 95 185 110,clip]{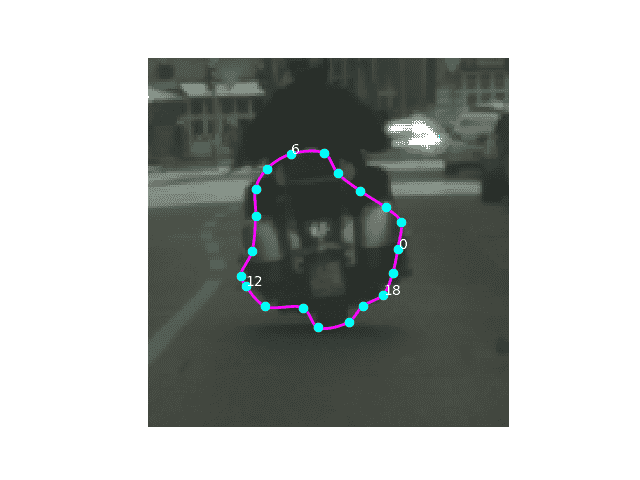}&
\includegraphics[height=1.85cm,trim=200 95 185 110,clip]{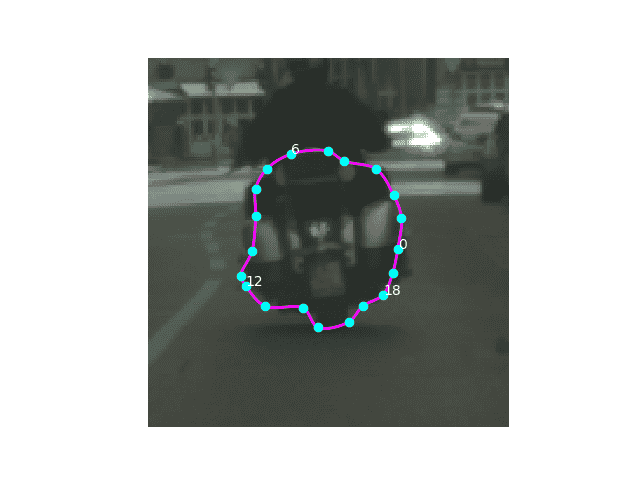} 
\end{tabular}
\vspace{-4.2mm}
\caption{\footnotesize {\bf Annotator in the Loop}: GT, 2nd column is the initial prediction from Spline-GCN, and the following columns show results after (simulated)  annotator's corrections. Our corrections are local, and thus give more control to the annotator. However, they sometimes require more clicks (right).}
\label{fig:human-in-loop-instance}
\vspace{-2mm}
\end{figure*}

% \paragraph{Pixel-wise Segmentation:} 
% We provide comparison with Deeplab~\cite{ChenPKMY18} in the Table ~\ref{tbl:F-seg}. Spline-GCN outperforms  Deeplab on IOU metric, and surpasses it significantly on boundary F score, which matters more when annotating instances.  
%Note that Deeplab~\cite{ChenPKMY18} doesn't allow annotator in loop. 
%We additionally compare Dextr-Spline-GCN with DEXTR~\cite{dextr} in terms of Boundary F score and Dextr-Spline-GCN is significantly better than DEXTR~\cite{dextr}. 
%We claim that our boundary segmentation method outperforms pixel-wise segmentation methods on IOU and is significantly better on boundary F score, 

\begin{table*}[t!]
\vspace{-2mm}
\begin{minipage}{0.46\linewidth}
\vspace{1.2mm}
\begin{center}
{\footnotesize
\addtolength{\tabcolsep}{-3pt}
\begin{tabular}{|l|c|c|c|c|c|}
\hline
Model &KITTI&ADE& Rooftop& Card.MR &ssTEM  \\
\hline
\hline
%SquareBox (Expansion)& - & 42.95 &40.71 &62.10& 42.24 \\
%Ellipse (Expansion) & - & 48.53 &47.51& 73.63& 51.04 \\
Square Box (Perfect) & - & 69.35 &62.11 &79.11 &66.53 \\
Ellipse (Perfect) & - &69.53& 66.82& 92.44 &71.32\\
\hline
\hline
Polygon-RNN++ (BS) & 83.14 & 71.82 & 65.67 & 80.63 &53.12 \\
\hline
PSP-DeepLab & 83.35 & 72.70 & 57.91  & 74.11  & 47.65 \\
% PSP-Deeplab & 83.35 / \textbf{78.05} & 72.70 / 46.15& 57.91 / 23.52 & 74.11 / 45.46 & 47.65 / 30.00 \\
% \hline
\hline
Spline-GCN & \textbf{84.09}  & \textbf{72.94} & \textbf{68.33} & 78.54  & 58.46   \\
\hspace{2mm}{+ finetune} & {\bf 84.81}   & 77.35 &  {\bf 78.21}  & {\bf 91.33}  & - \\
% \hspace{2mm}{poly-matching-finetune} & 83.97 / 77.28  & 77.35 / 52.69 &  70.16 / 33.14 & 78.72 / 55.16  & - \\
\hline
Polygon-GCN & 83.66   & 72.31 & 66.78 & \textbf{81.55}  & \textbf{60.91}  \\
\hspace{2mm}{+ finetune} & 84.71   & {\bf 77.41} &  75.56& 90.91  & - \\

% Spline-GCN & \textbf{84.09} / 76.32  & \textbf{72.94} / \textbf{47.09} & \textbf{68.33} / \textbf{29.60} & 78.54 / 55.25 & 58.46   / 45.09\\
% \hspace{2mm}{+ DiffAcc} & 84.81 / 77.28  & 77.35 / 52.69 &  78.21 / 41.86 & 91.33 / 91.50  & - \\
% % \hspace{2mm}{poly-matching-finetune} & 83.97 / 77.28  & 77.35 / 52.69 &  70.16 / 33.14 & 78.72 / 55.16  & - \\
% \hline
% Polygon-GCN & 83.66 / 75.84  & 72.31 / 46.67 & 66.78 / 26.96 & \textbf{81.55} / \textbf{60.64} & \textbf{60.91}   / \textbf{45.73}\\
% \hspace{2mm}{+ DiffAcc} & 84.71 / 77.29  & 77.41 / 52.36 &  75.56 / 39.17 & 90.91 / 89.67  & - \\
% \hspace{2mm}{poly-matching-finetune} & TBA  &TBA &  TBA & TBA  & - \\

% Deeplab & 83.35 / 78.05 & 72.70 / 46.15& 57.91 / 23.52 & 74.11 / 45.46 & 47.65 / 30.00 \\
% % \hline
% % Spline-GCN-multi-box & 84.11 / 76.92 & 73.00/ 47.95 & 72.11 / 32.89 & 84.18 / 70.47 & 47.83 / 38.46\\
% \hline
% Spline-GCN-Master & 84.09 / 76.32 & 72.94/ 47.09 & 68.33 / 29.60 & 78.54 / 55.25 & 58.46 / 45.09\\
% \hline
% Spline-GCN old & 84.07 / 77.35 & \textbf{73.34} / 48.82 /  &\textbf{69.33} / 33.28 & \textbf{78.31} / 55.98 & \textbf{58.48} / 47.20\\

% Spline-GCN New & 84.75 / 77.94 & \textbf{73.25} / 49.15 &55.37 / 28.61 & 77.70 / 55.71 & \textbf{53.18} / 45.87\\
% \hline \hline
% Dextr& 87.89/ 81.94 & 79.62 / 60.56 &81.67/ 48.35 & 88.97 / 79.54 & 73.81 / 64.81\\
% \hline
% Spline-GCN Dextr & 88.09 / 83.96 & 79.97 / 62.69 & 81.75 / 53.00 & 89.31 / 84.03& 68.16 / 64.35\\
\hline
\end{tabular}
\vspace{-3mm}
\caption{\footnotesize {\bf Automatic Mode on Cross-Domain}. We outperform PSP-DeepLab out-of-the-box. Fine-tuning on 10\% is effective. }
\label{tbl:cross-domain}
}
\end{center}
\end{minipage}
\hspace{1mm}
\begin{minipage}{0.53\linewidth}
\vspace{0.1mm}
\addtolength{\tabcolsep}{-5pt}
\begin{tabular}{ccccc}
\includegraphics[height=1.65cm,trim=12 30 5 0,clip]{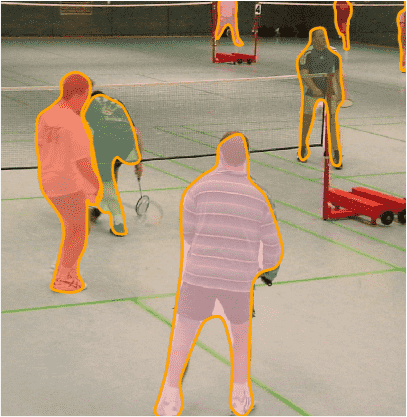} &
\includegraphics[height=1.65cm,trim=0 0 20 30,clip]{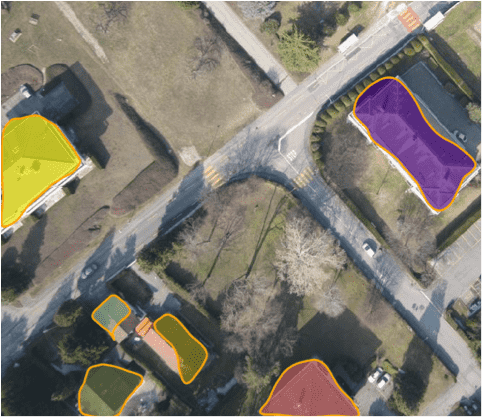} &
\includegraphics[height=1.65cm,trim=400 90 405 100,clip]{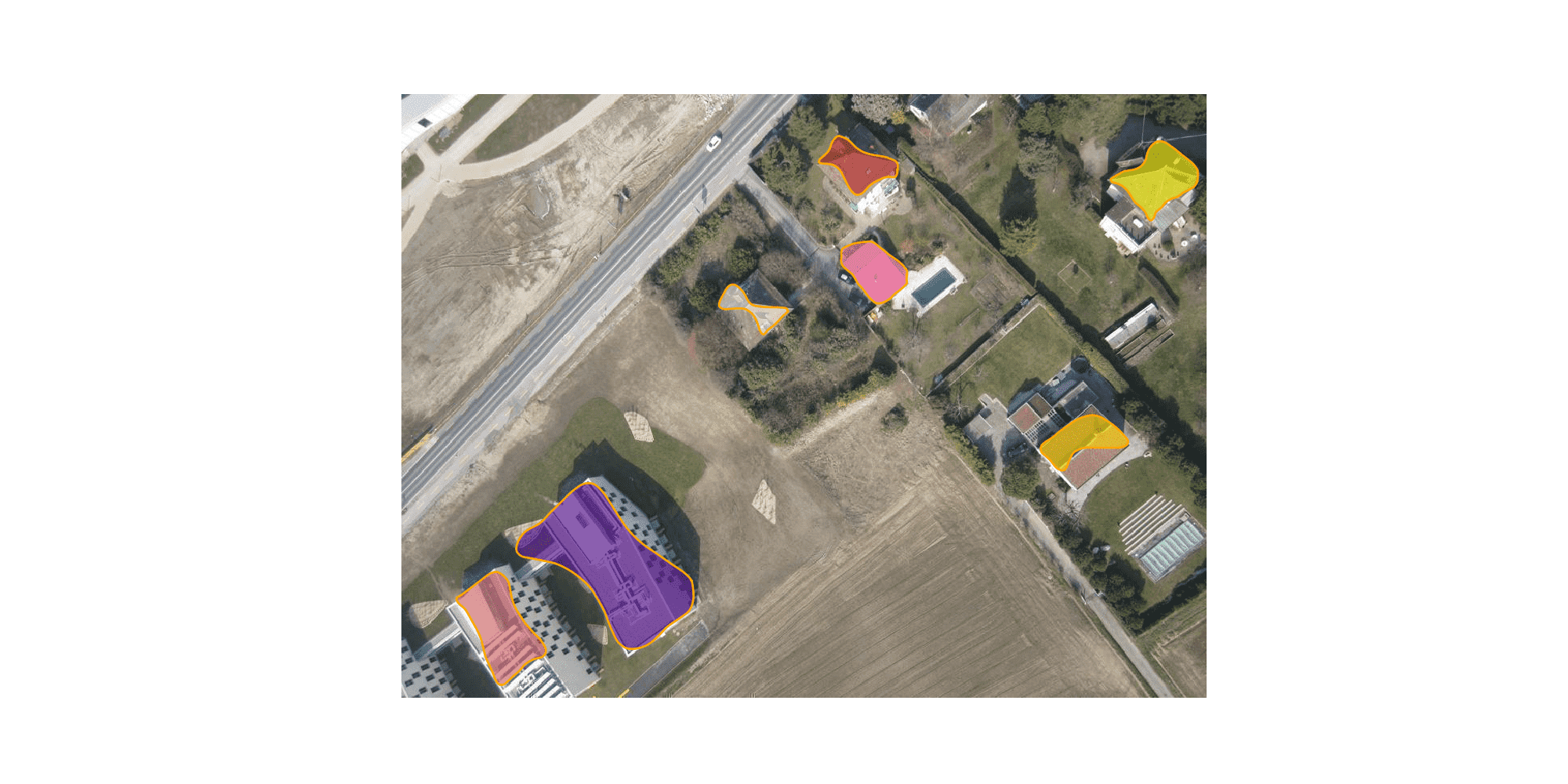} &
\includegraphics[height=1.65cm,trim=15 20 15 20,clip]{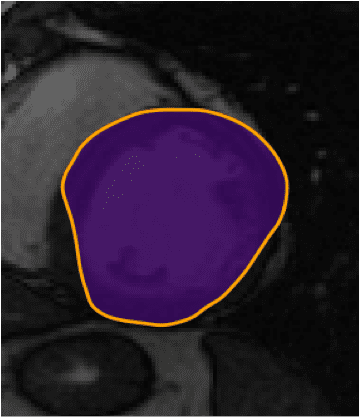} &
\includegraphics[height=1.65cm,trim=688 310 635 290,clip]{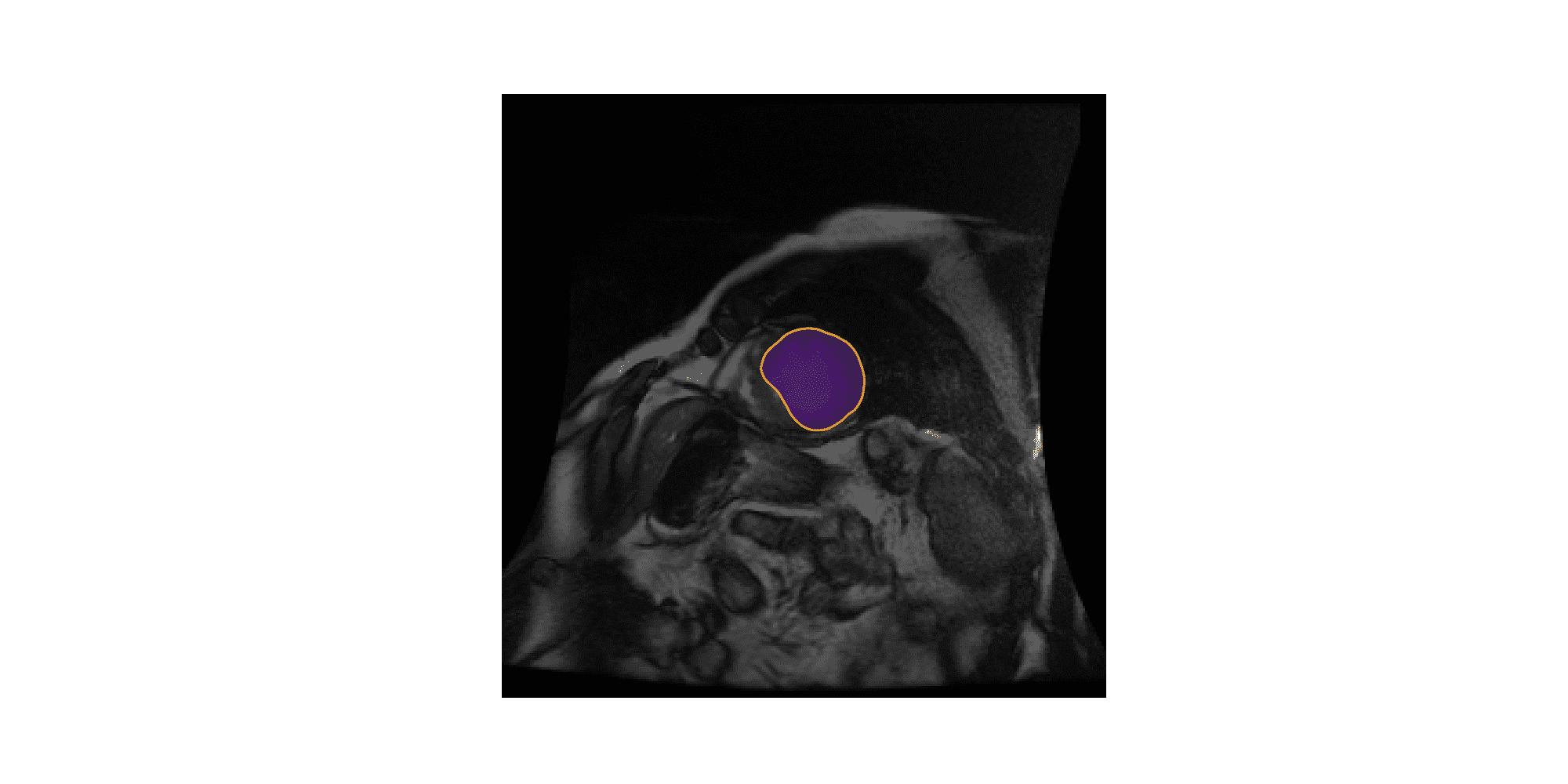} 
\\[-0.5mm]
\includegraphics[height=1.65cm,trim=7 30 0 0,clip]{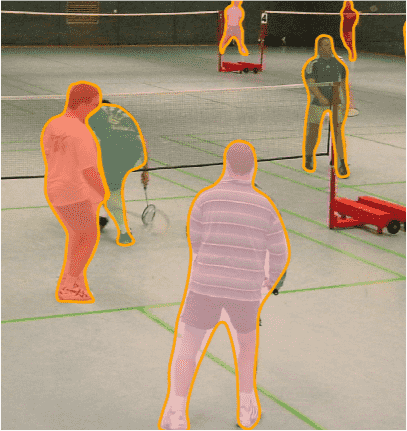} &
\includegraphics[height=1.65cm,trim=0 0 20 30,clip]{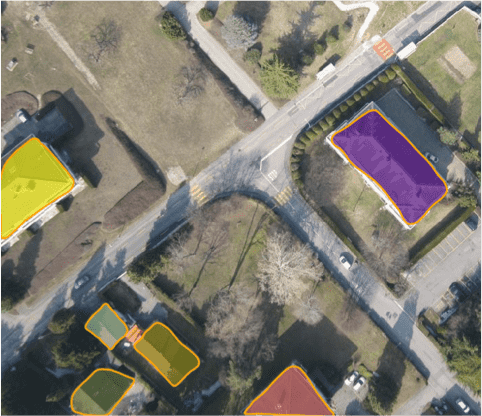} &
\includegraphics[height=1.65cm,trim=400 90 405 100,clip]{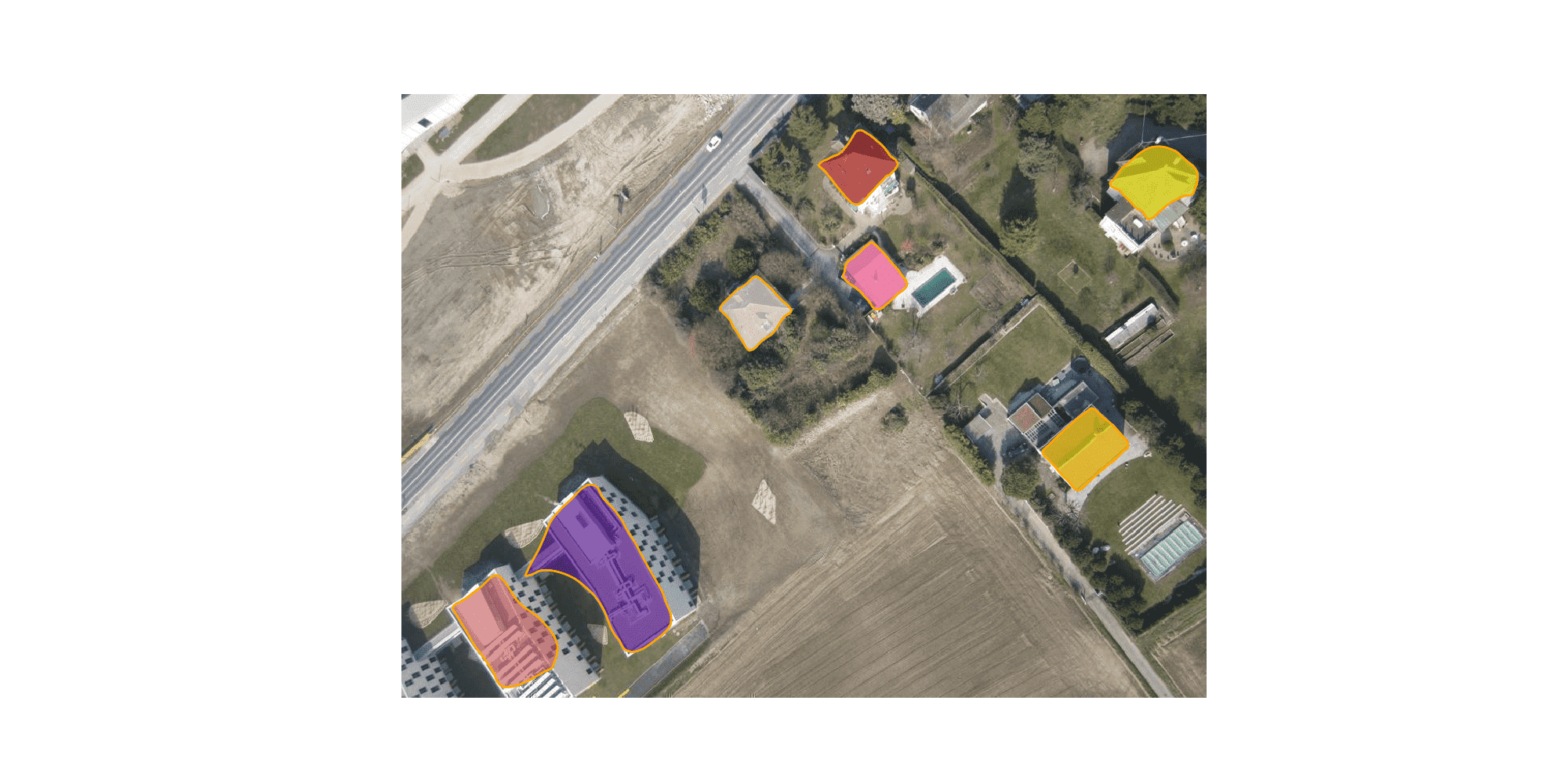} &
\includegraphics[height=1.65cm,trim=15 20 15 20,clip]{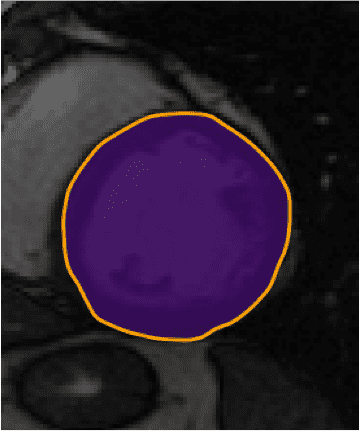} &
\includegraphics[height=1.65cm,trim=688 310 635 290,clip]{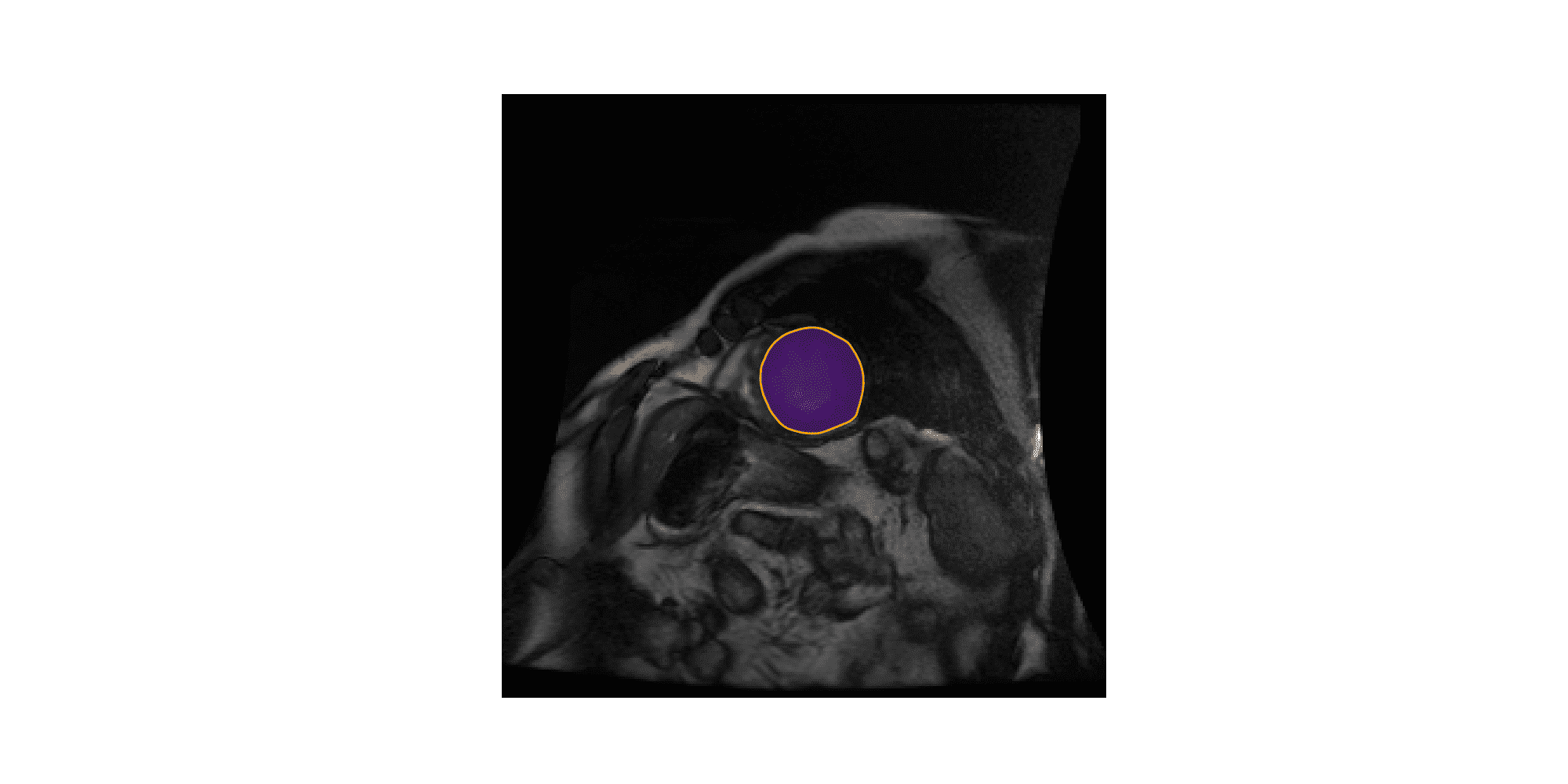}
\end{tabular}
\vspace{-4.5mm}
\caption{\footnotesize {\bf Automatic Mode for Cross-Domain}. ({\bf top}) Out-of-the-box output of Cityscapes-trained models, ({\bf bottom}) fine-tuned with 10\% of data from new domain.}
\label{fig:cross_domain_quantitive_results}
\vspace{-2mm}
\end{minipage}
\vspace{-5mm}
\end{table*}

\vspace{-4mm}
\paragraph{Interactive Mode:}
We simulate an annotator correcting vertices, following the protocol in~\cite{polyrnnpp}. In particular, the annotator iteratively makes corrections until the IoU is greater than a threshold $T$, or the model stops improving its prediction. We consider the predicted curve achieving agreement above $T$ as a satisfactory annotation. 
%\JUN{the meaning of T is not the same as polyrnn++. In polyrnn, for instance whose IOU > T, it won't correct and for other instance, it will run correction until the end. while ours could stop intermidiately. This is ours advantage.}

Plots~\ref{fig:human-in-loop} and ~\ref{fig:human-in-loop-kitti} show IoU vs number of clicks at different thresholds $T$. We compare to Polygon-RNN++. Our results show significant improvements over the baseline, highlighting our model as a more efficient annotation tool. We further analyze performance when using 40 vs 20 control points. The version with fewer control points is slightly worse in automatic mode (see Appendix), however, it is almost on par in the interactive mode. This may suggest that coarse-to-fine interactive correction may be the optimal approach. %\JUN{we remove the comparison on automatic mode for 40/20.}

\vspace{-6mm}
\paragraph{Inference Times:}  Timings are reported in Table~\ref{tbl:inference-time}. Our model is an order of magnitude faster than Polygon-RNN++, running at 29.3 ms, while Polygon-RNN++ requires 298.0ms. In the interactive mode, our model re-uses the computed image features computed in the forward pass, and thus only requires 2.6ms to incorporate each correction. On the other hand, Polygon-RNN requires to run an RNN after every correction, thus still requiring 270ms.

\subsection{Cross-Domain Evaluation}
\label{sec:cross-domain}

We now evaluate the ability of our model to generalize to new datasets. Generalization is crucial, in order to effectively  annotate a variety of different imagery types. We further show that by fine-tuning on only a small set of the new dataset (10\%) leads to fast adaptation to new domains. 
%In this section, we provide results of on cross domain datasets to evaluate the generalization capabilities of our model. We first show the results in which our model is directly tested to these datasets without any finetuning and then report results with finetuning. 

We follow~\cite{polyrnnpp} and use our Cityscapes-trained model and test it on KITTI~\cite{kitti} (in-domain driving dataset), ADE20k~\cite{ade20k} (general scenes), Rooftop~\cite{rooftop} (aerial imagery), and two medical datasets~\cite{cardiacmr,medical1,sstem}.  

\vspace{-4mm}
\paragraph{Quantitative Results.} Table~\ref{tbl:cross-domain} provides results. We adopt simple baselines from~\cite{polyrnnpp}. 
We further fine-tune (with diffAcc) the models with 10\% randomly sampled training data from the new domain. %We report numbers in Table~\ref{tbl:cross-domain} on all dataset except ssTEM  since ssTEM dataset doesn't have training set split. 
Note that ssTEM does not have a training split, and thus we omit this experiment for this dataset. 
Results show that our model generalizes better than PSP-DeepLab, and that fine-tuning on very little annotated data effectively adapts our model to new domains.  
Fig.~\ref{fig:cross_domain_quantitive_results} shows a few qualitative results before and after fine-tuning.

\vspace{-2mm}
\section{Conclusion}
\label{sec:conc}
\vspace{-0mm}

We presented Curve-GCN for efficient interactive annotation. Our model improves over the state-of-the-art and is significantly faster. We further allow interactive corrections that only have local effect, giving more control to the annotators. This leads to the better overall annotation strategy. We will release an annotation tool running our model, in order to facilitate faster collection of computer vision datasets.

{\small
\bibliographystyle{ieee}
\bibliography{egbib}
}

\end{document}

%%%%%%%%% TITLE
\title{\vspace{-15mm}Supplementary Material:\\ Fast Interactive Object Annotation with Curve-GCN}

\author{First Author\\
Institution1\\
Institution1 address\\
{\tt\small firstauthor@i1.org}
\and
Second Author\\
Institution2\\
First line of institution2 address\\
{\tt\small secondauthor@i2.org}
}

\maketitle

\fi

\section{Appendix}
%%%%%%%%% BODY TEXT
In the supplementary material, we provide additional qualitative and quantitative results for our Curve-GCN model. We compare models with different number of control points, to indicate the best effective topology. We further provide human-in-the-loop experiments for Polygon-GCN. Additional qualitative examples for both the automatic and interactive modes are given. Furthermore, we build a annotation tool and show several real human annotation examples.
\vspace{-2mm}
\subsection{Additional Quantitative Results}
\vspace{-2mm}
\paragraph{Ablating number of Control points.}
In Table~\ref{tbl:control_points}, we compare both Polygon-GCN and Spline-GCN trained with different numbers of control points. Both of them seem to follow a similar trend, with the sweet spot at 40 control points. This is also the number of control points used in most of the experiments in the main paper. Note however, that the differences between the models are not large, suggesting that for human-in-the-loop scenarios one may want to use fewer points instead. We show human-in-the-loop results for both 20 and 40 points in Figure~\ref{fig:poly-human-in-loop}.

\vspace{-4mm}
\paragraph{Interactive Polygon-GCN.} 
The results for Polygon-GCN in interactive mode are shown in the Figure~\ref{fig:poly-human-in-loop} and \ref{fig:poly-human-in-loop-kitti}. We compare our models with Polygon-RNN and Polygon-RNN++ on the Cityscapes dataset and KITTI. Notice that our model requires {\bf half} as many clicks than Polygon-RNN++, on average. 
\vspace{-2mm}
\subsection{Qualitative Results}
\vspace{-2mm}
\paragraph{Automatic Mode on Cityscapes.} 
In Figure~\ref{fig:human_auto_cityscapes} we provide a qualitative comparison between our model in automatic mode and the ground-truth polygons. We show our results in the first column and we emphasize that we exploit ground truth bounding boxes. Our model without any correction well matches annotators' work.
\vspace{-4mm}
\paragraph{Automatic Mode on Out-of-Domain Imagery.} 
We show out-of-domain predictions in Figures~\ref{fig:cross_domain1} and~\ref{fig:cross_domain2}. The first column shows ground truth results, while the second column shows Curve-GCN results. The third column depicts Curve-GCN results after fine-tuning on 10\% of training data. We also show IOU for each instance. While the out-of-the-box results are reasonable given that the model has never seen these objects or type of imagery, it quickly improves when fine-tuned on only a small portion of the new data.

\paragraph{Interactive Mode on Cityscapes.}
In Figure~\ref{fig:human-in-loop-instance-more}, we show qualitative results for the interactive mode with 40/20 control points in Cityscapes. We remind the reader that our correction is local, \ie when the annotator corrects a point only 2 of its neighbors on both the left and right side are predicted. Here $2$ is a hyperparameter, and other options are possible.

% \begin{table}[h!]
% \begin{center}
% {
% \addtolength{\tabcolsep}{2.5pt}
% \begin{tabular}{|l|c|c|c|c|c|}
% \hline
% \# control points & 20 pts & 30 pts & 40 pts & 50 pts & 60 pts \\
% \hline 
% \hline
% Poly-GCN &  70.06 & 70.55 & \textbf{71.19} & 71.11 & 71.06  \\
% Spline-GCN & 71.33 & 71.66 & \textbf{72.09} & 71.54 & 70.91  \\
% \hline
% \end{tabular}
% \caption{{\bf Automatic Mode on Cityscapes.} We compare IOU among models trained with different numbers of control points. } %Note that PSP-DeepLab uses a more powerful backbone architecture than all other models (which use Resnet). }
% \label{tbl:control_points}
% }
% \end{center}
% \end{table}

\begin{table}[h!]
\begin{center}
{
\addtolength{\tabcolsep}{2.5pt}
\begin{tabular}{|l|c|c|}
\hline
\ Mode & Poly-GCN & Spline-GCN  \\
\hline 
\hline 
20 pts & 70.06  & 71.33 \\              
30 pts & 70.55 & 71.66 \\
40 pts &  \textbf{71.19} & \textbf{72.09} \\
50 pts & 71.11 & 71.54 \\
60 pts & 71.06 & 70.91\\
\hline
\end{tabular}
\caption{{\bf Automatic Mode on Cityscapes.} We compare IOU among models trained with different numbers of control points. } %Note that PSP-DeepLab uses a more powerful backbone architecture than all other models (which use Resnet). }
\label{tbl:control_points}
}
\end{center}
\end{table}
\vspace{-4mm}
\subsection{Curve-GCN in a Real Annotation Tool}
 
We build a real time annotation tool based on our Curve-GCN model. Our tool allows annotators to annotate object instances manually or by using Curve-GCN in interactive mode. We show in-domain annotation results in Figure~\ref{fig:Real_tool_in_domain} and out-of-domain results in Figure~\ref{fig:Real_tool_out_domain}. The first column shows the image and bounding boxes, while the second column are results obtained when using interactive Curve-GCN. The third column is annotated in manual mode, where annotators are asked to label instances by drawing polygons. We also show the number of clicks for each mode. Our annotation tool will be released as well as hosted online.

\begin{figure}[h!]
	\centering
	\begin{minipage}{0.49\linewidth}
		\centering
		\includegraphics[height=2.7cm]{./figures/human_in_loop/city-40-fix-polygon.pdf}
		%\caption{Cityscapes Resnet 40 control points}
	\end{minipage}
	\begin{minipage}{0.49\linewidth}
		\centering
		\includegraphics[height=2.7cm]{./figures/human_in_loop/city-20-fix-polygon.pdf}
		%\caption{Cityscapes Resnet 20 control points}
	\end{minipage}
	% \vspace{-4mm}
	\caption{ {\bf Interactive Mode on Cityscapes with Polygon-GCN}: ({\bf left}) 40 control points, ({\bf right}) 20 control points.}
	\label{fig:poly-human-in-loop}
\end{figure}

\begin{figure}[h!]
	\centering
	\begin{minipage}{0.49\linewidth}
		\centering
		\includegraphics[height=2.7cm]{./figures/human_in_loop/kitti-40-fix-polygon.pdf}
		%\caption{Cityscapes Resnet 40 control points}
	\end{minipage}
	\begin{minipage}{0.49\linewidth}
		\centering
		\includegraphics[height=2.7cm]{./figures/human_in_loop/kitti-20-fix-polygon.pdf}
		%\caption{Cityscapes Resnet 20 control points}
	\end{minipage}
	% \vspace{-4mm}
	\caption{ {\bf Interactive Mode on KITTI with Polygon-GCN}: ({\bf left}) 40 control points, ({\bf right}) 20 control points.}
	\label{fig:poly-human-in-loop-kitti}
\end{figure}

\begin{figure*}[t!]
\addtolength{\tabcolsep}{-4.95pt}
\begin{tabular}{cccccccc}
\includegraphics[height=1.85cm,trim=180 95 145 105,clip]{figures/human_in_loop/resnet_40_points/frankfurt_000001_012699_70/frankfurt_000001_012699_70_stack_gt.png}  &
\includegraphics[height=1.85cm,trim=180 95 145 105,clip]{figures/human_in_loop/resnet_40_points/frankfurt_000001_012699_70/frankfurt_000001_012699_70_comp_0_click_0_stack_iou_74.png} &
\includegraphics[height=1.85cm,trim=180 95 145 105,clip]{figures/human_in_loop/resnet_40_points/frankfurt_000001_012699_70/frankfurt_000001_012699_70_comp_0_click_1_stack_iou_80.png} &
\includegraphics[height=1.85cm,trim=180 95 145 105,clip]{figures/human_in_loop/resnet_40_points/frankfurt_000001_012699_70/frankfurt_000001_012699_70_comp_0_click_2_stack_iou_83.png} &
\includegraphics[height=1.85cm,trim=180 95 145 105,clip]{figures/human_in_loop/resnet_40_points/frankfurt_000001_012699_70/frankfurt_000001_012699_70_comp_0_click_3_stack_iou_84.png} &
\includegraphics[height=1.85cm,trim=180 95 145 105,clip]{figures/human_in_loop/resnet_40_points/frankfurt_000001_012699_70/frankfurt_000001_012699_70_comp_0_click_4_stack_iou_84.png} &
\includegraphics[height=1.85cm,trim=180 95 145 105,clip]{figures/human_in_loop/resnet_40_points/frankfurt_000001_012699_70/frankfurt_000001_012699_70_comp_0_click_5_stack_iou_89.png} &
\includegraphics[height=1.85cm,trim=180 95 145 105,clip]{figures/human_in_loop/resnet_40_points/frankfurt_000001_012699_70/frankfurt_000001_012699_70_comp_0_click_6_stack_iou_88.png} \\
\includegraphics[height=1.85cm,trim=180 95 145 105,clip]{figures/human_in_loop/resnet_40_points/frankfurt_000001_054415_13/frankfurt_000001_054415_13_stack_gt.png} &
\includegraphics[height=1.85cm,trim=180 95 145 105,clip]{figures/human_in_loop/resnet_40_points/frankfurt_000001_054415_13/frankfurt_000001_054415_13_comp_0_click_0_stack_iou_55.png} &
\includegraphics[height=1.85cm,trim=180 95 145 105,clip]{figures/human_in_loop/resnet_40_points/frankfurt_000001_054415_13/frankfurt_000001_054415_13_comp_0_click_1_stack_iou_63.png} &
\includegraphics[height=1.85cm,trim=180 95 145 105,clip]{figures/human_in_loop/resnet_40_points/frankfurt_000001_054415_13/frankfurt_000001_054415_13_comp_0_click_2_stack_iou_71.png} &
\includegraphics[height=1.85cm,trim=180 95 145 105,clip]{figures/human_in_loop/resnet_40_points/frankfurt_000001_054415_13/frankfurt_000001_054415_13_comp_0_click_3_stack_iou_78.png} &
\includegraphics[height=1.85cm,trim=180 95 145 105,clip]{figures/human_in_loop/resnet_40_points/frankfurt_000001_054415_13/frankfurt_000001_054415_13_comp_0_click_4_stack_iou_85.png} &
\includegraphics[height=1.85cm,trim=180 95 145 105,clip]{figures/human_in_loop/resnet_40_points/frankfurt_000001_054415_13/frankfurt_000001_054415_13_comp_0_click_5_stack_iou_87.png} &
\includegraphics[height=1.85cm,trim=180 95 145 105,clip]{figures/human_in_loop/resnet_40_points/frankfurt_000001_054415_13/frankfurt_000001_054415_13_comp_0_click_6_stack_iou_92.png} \\
\includegraphics[height=1.85cm,trim=180 95 145 105,clip]{figures/human_in_loop/resnet_40_points/frankfurt_000001_068208_28/frankfurt_000001_068208_28_stack_gt.png} &
\includegraphics[height=1.85cm,trim=180 95 145 105,clip]{figures/human_in_loop/resnet_40_points/frankfurt_000001_068208_28/frankfurt_000001_068208_28_comp_0_click_0_stack_iou_72.png} &
\includegraphics[height=1.85cm,trim=180 95 145 105,clip]{figures/human_in_loop/resnet_40_points/frankfurt_000001_068208_28/frankfurt_000001_068208_28_comp_0_click_1_stack_iou_80.png} &
\includegraphics[height=1.85cm,trim=180 95 145 105,clip]{figures/human_in_loop/resnet_40_points/frankfurt_000001_068208_28/frankfurt_000001_068208_28_comp_0_click_2_stack_iou_84.png} &
\includegraphics[height=1.85cm,trim=180 95 145 105,clip]{figures/human_in_loop/resnet_40_points/frankfurt_000001_068208_28/frankfurt_000001_068208_28_comp_0_click_3_stack_iou_86.png} &
\includegraphics[height=1.85cm,trim=180 95 145 105,clip]{figures/human_in_loop/resnet_40_points/frankfurt_000001_068208_28/frankfurt_000001_068208_28_comp_0_click_4_stack_iou_88.png} &
\includegraphics[height=1.85cm,trim=180 95 145 105,clip]{figures/human_in_loop/resnet_40_points/frankfurt_000001_068208_28/frankfurt_000001_068208_28_comp_0_click_5_stack_iou_91.png} &
\includegraphics[height=1.85cm,trim=180 95 145 105,clip]{figures/human_in_loop/resnet_40_points/frankfurt_000001_068208_28/frankfurt_000001_068208_28_comp_0_click_6_stack_iou_93.png} \\
\includegraphics[height=1.85cm,trim=130 65 115 65,clip]{figures/human_in_loop/resnet_40_points/munster_000129_000019_104/munster_000129_000019_104_stack_gt.png}  &
\includegraphics[height=1.85cm,trim=130 65 115 65,clip]{figures/human_in_loop/resnet_40_points/munster_000129_000019_104/munster_000129_000019_104_comp_0_click_0_stack_iou_76.png} &
\includegraphics[height=1.85cm,trim=130 65 115 65,clip]{figures/human_in_loop/resnet_40_points/munster_000129_000019_104/munster_000129_000019_104_comp_0_click_1_stack_iou_82.png} &
\includegraphics[height=1.85cm,trim=130 65 115 65,clip]{figures/human_in_loop/resnet_40_points/munster_000129_000019_104/munster_000129_000019_104_comp_0_click_2_stack_iou_86.png} &
\includegraphics[height=1.85cm,trim=130 65 115 65,clip]{figures/human_in_loop/resnet_40_points/munster_000129_000019_104/munster_000129_000019_104_comp_0_click_3_stack_iou_88.png} &
\includegraphics[height=1.85cm,trim=130 65 115 65,clip]{figures/human_in_loop/resnet_40_points/munster_000129_000019_104/munster_000129_000019_104_comp_0_click_4_stack_iou_90.png} &
\includegraphics[height=1.85cm,trim=130 65 115 65,clip]{figures/human_in_loop/resnet_40_points/munster_000129_000019_104/munster_000129_000019_104_comp_0_click_5_stack_iou_90.png} &
\includegraphics[height=1.85cm,trim=130 65 115 65,clip]{figures/human_in_loop/resnet_40_points/munster_000129_000019_104/munster_000129_000019_104_comp_0_click_7_stack_iou_92.png} \\
\includegraphics[height=1.85cm,trim=180 95 145 105,clip]{figures/human_in_loop/resnet_20/frankfurt_000001_040575_89/frankfurt_000001_040575_89_stack_gt.png} &
\includegraphics[height=1.85cm,trim=180 95 145 105,clip]{figures/human_in_loop/resnet_20/frankfurt_000001_040575_89/frankfurt_000001_040575_89_comp_0_click_0_stack_iou_071.png} &
\includegraphics[height=1.85cm,trim=180 95 145 105,clip]{figures/human_in_loop/resnet_20/frankfurt_000001_040575_89/frankfurt_000001_040575_89_comp_0_click_1_stack_iou_082.png} &
\includegraphics[height=1.85cm,trim=180 95 145 105,clip]{figures/human_in_loop/resnet_20/frankfurt_000001_040575_89/frankfurt_000001_040575_89_comp_0_click_2_stack_iou_086.png} &
\includegraphics[height=1.85cm,trim=180 95 145 105,clip]{figures/human_in_loop/resnet_20/frankfurt_000001_040575_89/frankfurt_000001_040575_89_comp_0_click_3_stack_iou_086.png} &
\includegraphics[height=1.85cm,trim=180 95 145 105,clip]{figures/human_in_loop/resnet_20/frankfurt_000001_040575_89/frankfurt_000001_040575_89_comp_0_click_4_stack_iou_086.png} &
\includegraphics[height=1.85cm,trim=180 95 145 105,clip]{figures/human_in_loop/resnet_20/frankfurt_000001_040575_89/frankfurt_000001_040575_89_comp_0_click_5_stack_iou_090.png} &
\includegraphics[height=1.85cm,trim=180 95 145 105,clip]{figures/human_in_loop/resnet_20/frankfurt_000001_040575_89/frankfurt_000001_040575_89_comp_0_click_6_stack_iou_091.png} \\
\includegraphics[height=1.85cm,trim=180 95 145 105,clip]{figures/human_in_loop/add_resnet_20/frankfurt_000001_020046_80/frankfurt_000001_020046_80_stack_gt.png} &
\includegraphics[height=1.85cm,trim=180 95 145 105,clip]{figures/human_in_loop/add_resnet_20/frankfurt_000001_020046_80/frankfurt_000001_020046_80_comp_0_click_0_stack_iou_52.png} &
\includegraphics[height=1.85cm,trim=180 95 145 105,clip]{figures/human_in_loop/add_resnet_20/frankfurt_000001_020046_80/frankfurt_000001_020046_80_comp_0_click_1_stack_iou_77.png} &
\includegraphics[height=1.85cm,trim=180 95 145 105,clip]{figures/human_in_loop/add_resnet_20/frankfurt_000001_020046_80/frankfurt_000001_020046_80_comp_0_click_2_stack_iou_88.png} &
\includegraphics[height=1.85cm,trim=180 95 145 105,clip]{figures/human_in_loop/add_resnet_20/frankfurt_000001_020046_80/frankfurt_000001_020046_80_comp_0_click_3_stack_iou_88.png} &
\includegraphics[height=1.85cm,trim=180 95 145 105,clip]{figures/human_in_loop/add_resnet_20/frankfurt_000001_020046_80/frankfurt_000001_020046_80_comp_0_click_4_stack_iou_90.png} &
\includegraphics[height=1.85cm,trim=180 95 145 105,clip]{figures/human_in_loop/add_resnet_20/frankfurt_000001_020046_80/frankfurt_000001_020046_80_comp_0_click_4_stack_iou_90.png} &
\includegraphics[height=1.85cm,trim=180 95 145 105,clip]{figures/human_in_loop/add_resnet_20/frankfurt_000001_020046_80/frankfurt_000001_020046_80_comp_0_click_4_stack_iou_90.png} \\
\includegraphics[height=1.85cm,trim=140 65 115 75,clip]{figures/human_in_loop/add_resnet_20/frankfurt_000001_046779_91/frankfurt_000001_046779_91_stack_gt.png} &
\includegraphics[height=1.85cm,trim=140 65 115 75,clip]{figures/human_in_loop/add_resnet_20/frankfurt_000001_046779_91/frankfurt_000001_046779_91_comp_0_click_0_stack_iou_63.png} &
\includegraphics[height=1.85cm,trim=140 65 115 75,clip]{figures/human_in_loop/add_resnet_20/frankfurt_000001_046779_91/frankfurt_000001_046779_91_comp_0_click_1_stack_iou_89.png} &
\includegraphics[height=1.85cm,trim=140 65 115 75,clip]{figures/human_in_loop/add_resnet_20/frankfurt_000001_046779_91/frankfurt_000001_046779_91_comp_0_click_2_stack_iou_89.png} &
\includegraphics[height=1.85cm,trim=140 65 115 75,clip]{figures/human_in_loop/add_resnet_20/frankfurt_000001_046779_91/frankfurt_000001_046779_91_comp_0_click_3_stack_iou_89.png} &
\includegraphics[height=1.85cm,trim=140 65 115 75,clip]{figures/human_in_loop/add_resnet_20/frankfurt_000001_046779_91/frankfurt_000001_046779_91_comp_0_click_4_stack_iou_90.png} &
\includegraphics[height=1.85cm,trim=140 65 115 75,clip]{figures/human_in_loop/add_resnet_20/frankfurt_000001_046779_91/frankfurt_000001_046779_91_comp_0_click_5_stack_iou_90.png} &
\includegraphics[height=1.85cm,trim=140 65 115 75,clip]{figures/human_in_loop/add_resnet_20/frankfurt_000001_046779_91/frankfurt_000001_046779_91_comp_0_click_6_stack_iou_91.png} \\
\includegraphics[height=1.85cm,trim=140 65 115 75,clip]{figures/human_in_loop/add_resnet_20/munster_000009_000019_28/munster_000009_000019_28_stack_gt.png} &
\includegraphics[height=1.85cm,trim=140 65 115 75,clip]{figures/human_in_loop/add_resnet_20/munster_000009_000019_28/munster_000009_000019_28_comp_0_click_0_stack_iou_72.png} &
\includegraphics[height=1.85cm,trim=140 65 115 75,clip]{figures/human_in_loop/add_resnet_20/munster_000009_000019_28/munster_000009_000019_28_comp_0_click_1_stack_iou_89.png} &
\includegraphics[height=1.85cm,trim=140 65 115 75,clip]{figures/human_in_loop/add_resnet_20/munster_000009_000019_28/munster_000009_000019_28_comp_0_click_2_stack_iou_87.png} &
\includegraphics[height=1.85cm,trim=140 65 115 75,clip]{figures/human_in_loop/add_resnet_20/munster_000009_000019_28/munster_000009_000019_28_comp_0_click_3_stack_iou_89.png} &
\includegraphics[height=1.85cm,trim=140 65 115 75,clip]{figures/human_in_loop/add_resnet_20/munster_000009_000019_28/munster_000009_000019_28_comp_0_click_4_stack_iou_91.png} &
\includegraphics[height=1.85cm,trim=140 65 115 75,clip]{figures/human_in_loop/add_resnet_20/munster_000009_000019_28/munster_000009_000019_28_comp_0_click_5_stack_iou_92.png} &
\includegraphics[height=1.85cm,trim=140 65 115 75,clip]{figures/human_in_loop/add_resnet_20/munster_000009_000019_28/munster_000009_000019_28_comp_0_click_5_stack_iou_92.png} \\
\includegraphics[height=1.85cm,trim=140 65 115 75,clip]{figures/human_in_loop/add_resnet_20/munster_000094_000019_34/munster_000094_000019_34_stack_gt.png} &
\includegraphics[height=1.85cm,trim=140 65 115 75,clip]{figures/human_in_loop/add_resnet_20/munster_000094_000019_34/munster_000094_000019_34_comp_0_click_0_stack_iou_73.png} &
\includegraphics[height=1.85cm,trim=140 65 115 75,clip]{figures/human_in_loop/add_resnet_20/munster_000094_000019_34/munster_000094_000019_34_comp_0_click_1_stack_iou_80.png} &
\includegraphics[height=1.85cm,trim=140 65 115 75,clip]{figures/human_in_loop/add_resnet_20/munster_000094_000019_34/munster_000094_000019_34_comp_0_click_2_stack_iou_86.png} &
\includegraphics[height=1.85cm,trim=140 65 115 75,clip]{figures/human_in_loop/add_resnet_20/munster_000094_000019_34/munster_000094_000019_34_comp_0_click_3_stack_iou_88.png} &
\includegraphics[height=1.85cm,trim=140 65 115 75,clip]{figures/human_in_loop/add_resnet_20/munster_000094_000019_34/munster_000094_000019_34_comp_0_click_4_stack_iou_89.png} &
\includegraphics[height=1.85cm,trim=140 65 115 75,clip]{figures/human_in_loop/add_resnet_20/munster_000094_000019_34/munster_000094_000019_34_comp_0_click_5_stack_iou_93.png} &
\includegraphics[height=1.85cm,trim=140 65 115 75,clip]{figures/human_in_loop/add_resnet_20/munster_000094_000019_34/munster_000094_000019_34_comp_0_click_6_stack_iou_93.png} \\
\includegraphics[height=1.85cm,trim=140 65 115 75,clip]{figures/human_in_loop/add_resnet_20/munster_000088_000019_28/munster_000088_000019_28_stack_gt.png} &
\includegraphics[height=1.85cm,trim=140 65 115 75,clip]{figures/human_in_loop/add_resnet_20/munster_000088_000019_28/munster_000088_000019_28_comp_0_click_0_stack_iou_65.png} &
\includegraphics[height=1.85cm,trim=140 65 115 75,clip]{figures/human_in_loop/add_resnet_20/munster_000088_000019_28/munster_000088_000019_28_comp_0_click_1_stack_iou_82.png} &
\includegraphics[height=1.85cm,trim=140 65 115 75,clip]{figures/human_in_loop/add_resnet_20/munster_000088_000019_28/munster_000088_000019_28_comp_0_click_2_stack_iou_83.png} &
\includegraphics[height=1.85cm,trim=140 65 115 75,clip]{figures/human_in_loop/add_resnet_20/munster_000088_000019_28/munster_000088_000019_28_comp_0_click_3_stack_iou_88.png} &
\includegraphics[height=1.85cm,trim=140 65 115 75,clip]{figures/human_in_loop/add_resnet_20/munster_000088_000019_28/munster_000088_000019_28_comp_0_click_4_stack_iou_91.png} &
\includegraphics[height=1.85cm,trim=140 65 115 75,clip]{figures/human_in_loop/add_resnet_20/munster_000088_000019_28/munster_000088_000019_28_comp_0_click_5_stack_iou_90.png} &
\includegraphics[height=1.85cm,trim=140 65 115 75,clip]{figures/human_in_loop/add_resnet_20/munster_000088_000019_28/munster_000088_000019_28_comp_0_click_6_stack_iou_90.png} \\

\end{tabular}
% \vspace{-4.2mm}
\caption{ {\bf Annotator in the Loop}: The first column is GT, the 2nd column is the initial prediction from Spline-GCN, and the following columns show results after (simulated) annotator's corrections one by one.}
\label{fig:human-in-loop-instance-more}
% \vspace{-2mm}
\end{figure*}

\begin{figure*}[ht]
	\centering
	\begin{tabular}{c c}
	\bf{Curve GCN} & \bf{Human Annotator}\\[1mm]
	\includegraphics[width=0.496\linewidth,trim=180 200 150 190,clip]{figures/full_auto/frankfurt_000000_000576_iou_087_nclasses_19_labels_car_bicycle_rider_ncorrections_0} & \includegraphics[width=0.496\linewidth,trim=180 200 150 190,clip]{figures/full_auto/c_gt/frankfurt_000000_000576_iou_087_nclasses_19_ncorrections_0} \\[-0.3mm] 
	\bf{0 clicks} & \bf{700 clicks}\\[2mm]

	\includegraphics[width=0.496\linewidth,trim=180 280 150 100,clip]{figures/full_auto/frankfurt_000000_006589_iou_087_nclasses_9_labels_car_train_truck_ncorrections_0.png} & \includegraphics[width=0.496\linewidth,trim=180 280 150 100,clip]{figures/full_auto/c_gt/frankfurt_000000_006589_iou_087_nclasses_9_ncorrections_0}\\[-0.3mm] 
	\bf{0 clicks} & \bf{286 clicks}\\[2mm]

	\includegraphics[width=0.496\linewidth,trim=180 250 150 120,clip]{figures/full_auto/frankfurt_000000_011007_iou_088_nclasses_6_labels_car_truck_person_ncorrections_0.png} & \includegraphics[width=0.496\linewidth,trim=180 250 150 120,clip]{figures/full_auto/c_gt/frankfurt_000000_011007_iou_088_nclasses_6_ncorrections_0.png}\\[-0.3mm] 
	\bf{0 clicks} & \bf{161 clicks}\\[2mm]
	
	\includegraphics[width=0.496\linewidth,trim=180 180 150 200,clip]{figures/full_auto/frankfurt_000000_013067_iou_088_nclasses_11_labels_bus_person_bicycle_rider_car_ncorrections_0.png} & \includegraphics[width=0.496\linewidth,trim=180 180 150 200,clip]{figures/full_auto/c_gt/frankfurt_000000_013067_iou_088_nclasses_11_ncorrections_0.png}\\[-0.3mm] 
	\bf{0 clicks} & \bf{362 clicks}\\[2mm]
	
	\includegraphics[width=0.496\linewidth,trim=180 170 150 220,clip]{figures/full_auto/munster_000035_000019_iou_091_nclasses_10_labels_car_person_ncorrections_0.png} & \includegraphics[width=0.496\linewidth,trim=180 170 150 220,clip]{figures/full_auto/c_gt/munster_000035_000019_iou_091_nclasses_10_ncorrections_0.png}\\[-0.3mm] 
	\bf{0 clicks} & \bf{435 clicks}\\[2mm]

	\includegraphics[width=0.496\linewidth,trim=180 140 150 250,clip]{figures/full_auto/munster_000082_000019_iou_092_nclasses_12_labels_car_bicycle_ncorrections_0} & \includegraphics[width=0.496\linewidth,trim=180 140 150 250,clip]{figures/full_auto/c_gt/munster_000082_000019_iou_092_nclasses_12_ncorrections_0}\\[-0.3mm] 
	\bf{0 clicks} & \bf{554 clicks}\\
\end{tabular}
\caption{Qualitative comparison between Curve-GCN in automatic mode and ground-truth, on Cityscapes. Note that our model relies on bounding boxes.}
\label{fig:human_auto_cityscapes}
\end{figure*}

\begin{figure*}[ht]
	\centering
	\begin{tabular}{c c}
	\includegraphics[width=0.496\linewidth,trim=100 0 100 0,clip]{figures/cross_domain/roof/IMG_0514_0_iou_048_iou_after_087} & 
	\includegraphics[width=0.496\linewidth,trim=40 0 40 0,clip]{figures/cross_domain/roof/IMG_0516_6_iou_076_iou_after_087} \\[-0.3mm] 
	\bf{Before: IOU 0.48, After: IOU 0.87} & \bf{Before: IOU 0.76, After: IOU 0.87}\\[2mm]

	\includegraphics[width=0.496\linewidth,trim=100 0 100 0,clip]{figures/cross_domain/roof/IMG_0572_3_iou_083_iou_after_092} & 
	\includegraphics[width=0.496\linewidth,trim=40 0 40 0,clip]{figures/cross_domain/roof/IMG_0574_5_iou_065_iou_after_089} \\[-0.3mm] 
	\bf{Before: IOU 0.83, After: IOU 0.92} & \bf{Before: IOU 0.65, After: IOU 0.89}\\[2mm]

	\includegraphics[width=0.496\linewidth,trim=100 0 100 0,clip]{figures/cross_domain/roof/IMG_0595_1_iou_088_iou_after_090} & 
	\includegraphics[width=0.496\linewidth,trim=40 0 40 0,clip]{figures/cross_domain/roof/IMG_0693_8_iou_056_iou_after_091} \\[-0.3mm] 
	\bf{Before: IOU 0.88, After: IOU 0.90} & \bf{Before: IOU 0.56, After: IOU 0.91}\\[2mm]

	\includegraphics[width=0.496\linewidth,trim=100 0 100 0,clip]{figures/cross_domain/kitti/germany_000001_000779_0_iou_093_iou_after_095} & 
	\includegraphics[width=0.496\linewidth,trim=40 0 40 0,clip]{figures/cross_domain/kitti/germany_000001_000970_0_iou_095_iou_after_096} \\[-0.3mm] 
	\bf{Before: IOU 0.93, After: IOU 0.95} & \bf{Before: IOU 0.95, After: IOU 0.96}\\[2mm]

	\includegraphics[width=0.496\linewidth,trim=100 0 100 0,clip]{figures/cross_domain/kitti/germany_000001_001136_5_iou_091_iou_after_092} & 
	\includegraphics[width=0.496\linewidth,trim=40 0 40 0,clip]{figures/cross_domain/kitti/germany_000001_001411_2_iou_093_iou_after_096} \\[-0.3mm] 
	\bf{Before: IOU 0.91, After: IOU 0.92} & \bf{Before: IOU 0.93, After: IOU 0.96}\\[2mm]

	\includegraphics[width=0.496\linewidth,trim=100 0 100 0,clip]{figures/cross_domain/kitti/germany_000001_002964_0_iou_095_iou_after_096} & 
	\includegraphics[width=0.496\linewidth,trim=40 0 40 0,clip]{figures/cross_domain/kitti/germany_000001_003187_1_iou_089_iou_after_092} \\[-0.3mm] 
	\bf{Before: IOU 0.95, After: IOU 0.96} & \bf{Before: IOU 0.89, After: IOU 0.92}\\[2mm]

\end{tabular}
\caption{Cross domain qualitative comparison between a human annotator vs Curve-GCN and Curve-GCN fine-tuned on 10\% of training data.}
\label{fig:cross_domain1}
\end{figure*}

\begin{figure*}[ht]
	\centering
	\begin{tabular}{c c}
	\includegraphics[width=0.496\linewidth,trim=100 0 100 0,clip]{figures/cross_domain/medical/SUNNYBROOK-SC-HF-I-5-IM-80_0_iou_075_iou_after_094} & 
	\includegraphics[width=0.496\linewidth,trim=40 0 40 0,clip]{figures/cross_domain/medical/SUNNYBROOK-SC-HF-I-5-IM-200_0_iou_077_iou_after_089} \\[-0.3mm] 
	\bf{Before: IOU 0.75, After: IOU 0.94} & \bf{Before: IOU 0.77, After: IOU 0.89}\\[2mm]

	\includegraphics[width=0.496\linewidth,trim=100 0 100 0,clip]{figures/cross_domain/medical/SUNNYBROOK-SC-HF-I-8-IM-60_0_iou_079_iou_after_091} & 
	\includegraphics[width=0.496\linewidth,trim=40 0 40 0,clip]{figures/cross_domain/medical/SUNNYBROOK-SC-HF-NI-33-IM-120_0_iou_076_iou_after_093} \\[-0.3mm] 
	\bf{Before: IOU 0.79, After: IOU 0.91} & \bf{Before: IOU 0.76, After: IOU 0.93}\\[2mm]

	\includegraphics[width=0.496\linewidth,trim=100 0 100 0,clip]{figures/cross_domain/medical/SUNNYBROOK-SC-HYP-7-IM-160_0_iou_084_iou_after_090} & 
	\includegraphics[width=0.496\linewidth,trim=40 0 40 0,clip]{figures/cross_domain/medical/SUNNYBROOK-SC-HYP-8-IM-120_0_iou_088_iou_after_092} \\[-0.3mm] 
	\bf{Before: IOU 0.84, After: IOU 0.90} & \bf{Before: IOU 0.88, After: IOU 0.92}\\[2mm]

	\includegraphics[width=0.496\linewidth,trim=100 0 100 0,clip]{figures/cross_domain/ade/ADE_val_00001724_27_iou_080_iou_after_091} & 
	\includegraphics[width=0.496\linewidth,trim=40 0 40 0,clip]{figures/cross_domain/ade/ADE_val_00001813_1_iou_058_iou_after_080} \\[-0.3mm] 
	\bf{Before: IOU 0.80, After: IOU 0.91} & \bf{Before: IOU 0.58, After: IOU 0.80}\\[2mm]

	\includegraphics[width=0.496\linewidth,trim=100 0 100 0,clip]{figures/cross_domain/ade/ADE_val_00000229_18_iou_047_iou_after_082} & 
	\includegraphics[width=0.496\linewidth,trim=40 0 40 0,clip]{figures/cross_domain/ade/ADE_val_00000819_8_iou_092_iou_after_094} \\[-0.3mm] 
	\bf{Before: IOU 0.47, After: IOU 0.82} & \bf{Before: IOU 0.92, After: IOU 0.94}\\[2mm]

	\includegraphics[width=0.496\linewidth,trim=100 0 100 0,clip]{figures/cross_domain/ade/ADE_val_00001029_7_iou_086_iou_after_091} & 
	\includegraphics[width=0.496\linewidth,trim=40 0 40 0,clip]{figures/cross_domain/ade/ADE_val_00001376_39_iou_092_iou_after_091} \\[-0.3mm] 
	\bf{Before: IOU 0.86, After: IOU 0.91} & \bf{Before: IOU 0.92, After: IOU 0.91}\\[2mm]

\end{tabular}
\caption{Cross domain qualitative comparison between a human annotator vs Curve-GCN and Curve-GCN fine-tuned on 10\% of training data.}
\label{fig:cross_domain2}
\end{figure*}

\begin{figure*}[ht]
	\centering
	\begin{tabular}{c c c}
	\bf{Image} & \bf{Interactive Curve GCN} & \bf{Manual Annotation}\\[1mm]
	\includegraphics[width=0.3\linewidth,trim=0 0 0 0,clip]{figures/real_tool/1_img} & 
	\includegraphics[width=0.3\linewidth,trim=0 0 0 0,clip]{figures/real_tool/1_auto_0} &
	\includegraphics[width=0.3\linewidth,trim=0 0 0 0,clip]{figures/real_tool/1_gt_28}\\[-0.3mm] 
	\bf{-} & \bf{0 clicks} & \bf{28 clicks}\\[2mm]

	\includegraphics[width=0.3\linewidth,trim=0 0 0 0,clip]{figures/real_tool/2_img} & 
	\includegraphics[width=0.3\linewidth,trim=0 0 0 0,clip]{figures/real_tool/2_auto_1} &
	\includegraphics[width=0.3\linewidth,trim=0 0 0 0,clip]{figures/real_tool/2_gt_20}\\[-0.3mm] 
	\bf{-} & \bf{1 clicks} & \bf{20 clicks}\\[2mm]

	\includegraphics[width=0.3\linewidth,trim=0 0 0 0,clip]{figures/real_tool/3_img} & 
	\includegraphics[width=0.3\linewidth,trim=0 0 0 0,clip]{figures/real_tool/3_auto_0} &
	\includegraphics[width=0.3\linewidth,trim=0 0 0 0,clip]{figures/real_tool/3_gt_32}\\[-0.3mm] 
	\bf{-} & \bf{0 clicks} & \bf{32 clicks}\\[2mm]

	\includegraphics[width=0.3\linewidth,trim=0 0 0 0,clip]{figures/real_tool/4_img} & 
	\includegraphics[width=0.3\linewidth,trim=0 0 0 0,clip]{figures/real_tool/4_auto_1} &
	\includegraphics[width=0.3\linewidth,trim=0 0 0 0,clip]{figures/real_tool/4_gt_30}\\[-0.3mm] 
	\bf{-} & \bf{1 clicks} & \bf{30 clicks}\\[2mm]

		\includegraphics[width=0.3\linewidth,trim=0 0 0 0,clip]{figures/real_tool/12_img} & 
	\includegraphics[width=0.3\linewidth,trim=0 0 0 0,clip]{figures/real_tool/12_auto_1.png} &
	\includegraphics[width=0.3\linewidth,trim=0 0 0 0,clip]{figures/real_tool/12_gt_28}\\[-0.3mm] 
	\bf{-} & \bf{1 clicks} & \bf{28 clicks}\\[2mm]

			\includegraphics[width=0.3\linewidth,trim=0 0 0 0,clip]{figures/real_tool/13_img} & 
	\includegraphics[width=0.3\linewidth,trim=0 0 0 0,clip]{figures/real_tool/13_auto_2} &
	\includegraphics[width=0.3\linewidth,trim=0 0 0 0,clip]{figures/real_tool/13_gt_32}\\[-0.3mm] 
	\bf{-} & \bf{2 clicks} & \bf{32 clicks}\\[2mm]
	% \includegraphics[width=0.496\linewidth,trim=180 200 150 180,clip]{figures/full_auto/munster_000118_000019_iou_0_86} & \includegraphics[width=0.496\linewidth,trim=180 200 150 180,clip]{figures/full_auto/c_gt/munster_000118_000019_gt_clicks_245}\\
	% \bf{0 clicks} & \bf{245 clicks}\\

\end{tabular}
\caption{{\bf Our annotation tool:} Real tool annotation results on in-domain images.}
\label{fig:Real_tool_in_domain}
\end{figure*}

\begin{figure*}[ht]
	\centering
	\begin{tabular}{c c c}
	\bf{Image} & \bf{Interactive Curve GCN} & \bf{Manual Annotation}\\[1mm]
	\includegraphics[width=0.3\linewidth,trim=0 0 0 0,clip]{figures/real_tool/5_img.png} & 
	\includegraphics[width=0.3\linewidth,trim=0 0 0 0,clip]{figures/real_tool/5_auto_5} &
	\includegraphics[width=0.3\linewidth,trim=0 0 0 0,clip]{figures/real_tool/5_gt_40}\\[-0.3mm] 
	\bf{-} & \bf{5 clicks} & \bf{40 clicks}\\[2mm]

	\includegraphics[width=0.3\linewidth,trim=0 0 0 70,clip]{figures/real_tool/6_img} & 
	\includegraphics[width=0.3\linewidth,trim=0 0 0 70,clip]{figures/real_tool/6_auto_1} &
	\includegraphics[width=0.3\linewidth,trim=0 0 0 70,clip]{figures/real_tool/6_gt_26}\\[-0.3mm] 
	\bf{-} & \bf{1 clicks} & \bf{26 clicks}\\[2mm]

	\includegraphics[width=0.3\linewidth,trim=0 0 0 0,clip]{figures/real_tool/7_img} & 
	\includegraphics[width=0.3\linewidth,trim=0 0 0 0,clip]{figures/real_tool/7_auto_0} &
	\includegraphics[width=0.3\linewidth,trim=0 0 0 0,clip]{figures/real_tool/7_gt_37}\\[-0.3mm] 
	\bf{-} & \bf{0 clicks} & \bf{37 clicks}\\[2mm]

\includegraphics[width=0.3\linewidth,trim=0 0 0 0,clip]{figures/real_tool/11_img} & 
	\includegraphics[width=0.3\linewidth,trim=0 0 0 0,clip]{figures/real_tool/11_auto_1} &
	\includegraphics[width=0.3\linewidth,trim=0 0 0 0,clip]{figures/real_tool/11_gt_22}\\[-0.3mm] 
	\bf{-} & \bf{1 clicks} & \bf{22 clicks}\\[2mm]

	\includegraphics[width=0.3\linewidth,trim=0 0 0 0,clip]{figures/real_tool/10_img} & 
	\includegraphics[width=0.3\linewidth,trim=0 0 0 0,clip]{figures/real_tool/10_auto_7} &
	\includegraphics[width=0.3\linewidth,trim=0 0 0 0,clip]{figures/real_tool/10_gt_46}\\[-0.3mm] 
	\bf{-} & \bf{7 clicks} & \bf{46 clicks}\\[2mm]
	% \includegraphics[width=0.496\linewidth,trim=180 200 150 180,clip]{figures/full_auto/munster_000118_000019_iou_0_86} & \includegraphics[width=0.496\linewidth,trim=180 200 150 180,clip]{figures/full_auto/c_gt/munster_000118_000019_gt_clicks_245}\\
	% \bf{0 clicks} & \bf{245 clicks}\\

\end{tabular}
\caption{{\bf Our annotation tool:} Real tool annotation results on out-of-domain images.}
\label{fig:Real_tool_out_domain}
\end{figure*}

\iffalse
\clearpage
{\small
\bibliographystyle{ieee}
\bibliography{egbib}
}